\documentclass{article}

\usepackage{arxiv2}

\usepackage[utf8]{inputenc} 
\usepackage[T1]{fontenc}    
\usepackage{hyperref}       
\usepackage{url}            
\usepackage{booktabs}       
\usepackage{amsfonts}       
\usepackage{nicefrac}       
\usepackage{microtype}      
\usepackage{lipsum}		
\usepackage{multicol}
\usepackage{xcolor}
\usepackage{amsmath}
\usepackage{graphicx}
\usepackage{subfig}
\usepackage{longtable}
\usepackage[normalem]{ulem}

\title{A Hybrid Objective Function for Robustness of Artificial Neural Networks - Estimation of Parameters in a Mechanical System}

\author{
  Jan ~Sokolowski\\
  Department of Mathematics\\
  Trier University\\
  Department of Basics \& Mathematical Models\\
  IEE S.A. Bissen\\
  \texttt{jan.sokolowski@iee.lu} \\
   \And
  Volker ~Schulz \\
  Department of Mathematics\\
  Trier University\\
  \texttt{volker.schulz@uni-trier.de} \\  
  \And
  Udo ~Schr\"oder\\
  Department of Basics \& Mathematical Models\\
  IEE S.A. Bissen\\
  \texttt{udo.schroeder@iee.lu} \\  
  \And
  Hans - Peter ~Beise\\
  Department of Computer Science\\
  Trier University of Applied Sciences\\
  \texttt{h.beise@hochschule-trier.de} \\ 
}

\begin{document}
\maketitle

\begin{abstract}
In several studies, hybrid neural networks have proven to be more robust against noisy input data compared to plain data driven neural networks. 
We consider the task of estimating parameters of a mechanical vehicle model based on acceleration profiles. We introduce a convolutional neural network architecture that is capable to predict the parameters for a family of vehicle models that differ in the unknown parameters.
We introduce a convolutional neural network architecture that given sequential data predicts the parameters of the underlying data's dynamics.
This network is trained with two objective functions. The first one constitutes a more naive approach that assumes that the true parameters are known. The second objective incorporates the knowledge of the underlying dynamics and is therefore considered as hybrid approach. We show that in terms of robustness, the latter outperforms the first objective on noisy input data.
\end{abstract}


\section{Introduction}

Physical, biological and chemical models are important tools in nearly all fields of the engineering disciplines. In many cases, a major barrier for their application in practical cases is the lack of sufficient knowledge about the actual parameters \cite{Hamilton2017}.\\
As a consequence, a lot of technical knowledge in terms of well established equations cannot be exploit. The resulting problems to determine unknown parameters is a typical case of \textit{System Identification} \cite{Keesman2011, Brunton2016}.\\
In this work, we consider the case that a physical model is given as a dynamical system in terms of a differential equation. More specifically, we address the challenge of identification of parameters in a mechanical vehicle model, that is a coupled mass-spring-damper system \cite{Kumar2018}.\\
The practical application that we are targeting at is as follows. Given a family of such models, we assume that only the parameters that scale with the occupant's mass are unknown and the remaining parameters are fixed. We want to create a common model that predicts the unknown parameters based on acceleration profiles induced by realistic but randomly generated road profiles.\\
Deep neural networks have proven to bear great potential in many complex tasks. The remarkable improvements in computer vision \cite{He2015, Krizhevsky2012, Szedegy2015} and natural language processing \cite{Devlin2019, Mikolov2013} are undoubtedly among the most famous achievements in this context. In the course of this progress, deep neural networks have successfully applied in various other disciplines like reinforcement learning \cite{Mnih2013} or practical applications in health informatics \cite{Ravi2016} and energy consumption prediction \cite{Li2017}.\\
In a considerable line of publications, researchers attempted to tackle problems emerging in the context of physical equations. Mostly, data driven approaches can be used to generate approximative functions of a PDE structure \cite{Rudy2018} or to simulate the dynamical behaviour of time-dependent ODEs \cite{Qin2019, Rudy2019}. An appropriate network structure that fits the nature of the considered problem is often the key to successful results \cite{Chen2019, Yue2018}.\\
A common challenge that occurs in natural way in those works, is the open question of how to combine neural networks with prior knowledge. That is here, the understanding of the underlying physical laws \cite{Karpatne2018, Psichogios1992, Raissi2017p1, Raissi2017p2} that drives the data structure of the neural network's input.\\
In the present work, we use convolutional neural networks to process acceleration profiles. In that respect, we follow the works \cite{Yang2015, Zeng2014}. By doing so, we show that, based on simulated data, this kind of network can predict the unknown parameters. Our work is mostly related to \cite{Assidjo2009, Ayed2019, Rahim2016}.\\
This work provides the following contributions:\\
We compare the performance of one neural network for two different optimization processes. For both training processes, the target is to approximate a subset of the parameters of a system matrix that describes a set of ordinary differential equations. The first naive approach uses the true coefficients of the system matrix as labels, the second one recomputes the input data to indirectly approximate the parameters that are hidden within the data. We observe improved robustness against noisy test samples when using the second approach for neural network training.\\
\\
The paper is organised as follows: We discuss a methodology to compute an appropriate dataset that can be used to (partially) identify the system matrix of the underlying differential equations in Section $2$. Therefore, we describe, how to model the displacement of the road (Section 2.1) that can be used to compute the displacement of a passenger in an approximative vehicle model (Section 2.2), mathematically defined via a system of second order differential equations. Structure-preserving numerical algorithms like semi-implicit Euler methods (Section 2.3) can then help to generate synthetic datasets, consisting of the discrete acceleration profiles for the system of second order ODE. Then this sequential data can be used and processed by convolutional neural networks (Section 2.4), using multiple input channels to achieve good approximations of the true system parameters in the output layer. The concept of Section 2 is validated in Section 3, using a neural network to predict the missing parameters that are necessary to describe the acceleration $\ddot{z}$. We can further assume that the true parameters are known for the training process (Section 3.1) as a labelled learning approach and compare it to unlabelled learning (Section 3.2). For the unlabelled approach, the output of the neural network is used to reproduce the acceleration $\ddot{z}$. Then the true parameter values should result from minimizing the distance of true and reproduced acceleration. We can compare the performance of these two objectives for clean training and test data to clean training and noisy test data (Section 3.3). Finally, we can draw a conclusion, when to prefer a labelled or an unlabelled approach with prior knowledge in Section 4.\\

\section{Methodology}
We use a general non-homogeneous system of ordinary differential equations, mathematically defined by 
\begin{equation}
\dot{\rho}(t) = A \cdot \rho(t) + f(t),
\end{equation} 
where $\rho \in \mathcal{C}^1([t_0, t_N], \mathbb{R}^d)$, with $N, d \in \mathbb{N}$, is a multi-dimensional time-dependent state variable and the derivative with respect to time is denoted by $\dot{\rho}(t) = \frac{\mathrm{d}\rho(t)}{\mathrm{d}t}$, an exterior term$f \in \mathcal{C}([t_0, t_N], \mathbb{R}^d)$ that reacts on the dynamical system and $A \in \mathbb{R}^{d \times d}$ the so called system matrix with $\mathbb{T} := [t_0, t_N] \subset [0, \infty)$. We therefore discuss in this section, how to model a dynamical system of coupled rigid bodies, as mathematically described by Eq.$(1)$ in order to develop appropriate data and a sufficient neural network architecture for robust parameter estimation \cite{Hamilton2011, Peifer2007, Raol1996} for parts of the system matrix.\\ 
Appropriate data samples that represent a realisation of Eq.$(1)$ for varying system matrices $A$ can be computed by numerically solving the differential equation.\\
Therefore, we choose a physical model that can be accurately described by a simple system of second order ODEs. This system of second order ODEs can easily be reduced to a first order ODE system, which can in general be described using a representation as given by Eq.$(1)$. Using a structure-preserving numerical solution algorithm then results in data samples with a dynamical behaviour described by the underlying ODEs.\\
In this context, we use acceleration that can be physically described in terms of state and velocity with constant coefficients.

\subsection{Road Modelling}
Before describing the dynamical system itself, we discuss in detail, how an appropriate non-homogeneous term $f(t)$ with $t \in [t_0, t_N]$ can be derived in a meaningful way for a mechanical system. The following description of modelling road profiles complient to ISO 8606 standard is based on \cite{Tyan2009}.\\
Assume the distance, a car reaches on a specific road with absolute constant velocity, can be described by the continuous interval $\mathbb{S} := [s_0, s_N] \subset [0, \infty)$, where $s_N$ is the maximum distance with respect to the starting point $s_0$.
We define the state $r: \mathbb{S} \rightarrow \mathbb{R}$ that describes the displacement of the road at a point $s$ via:
\begin{equation}
r(s) = \sum_{i=1}^M A_i \sin(\omega_i s - \varphi_i),
\end{equation}
where $M \in \mathbb{N}$ is the number of relevant frequencies and the amplitude
\begin{equation}
A_i = \sqrt{\Phi(\omega_i)\left(\frac{\Delta \omega}{\pi}\right)}, \hspace{1mm} i = 1, 2, \hdots, M
\end{equation}
depends on the degree of roughness
\begin{equation}
\Phi(\omega_i) = \Phi(\omega_0)\left(\frac{\omega_i}{\omega_0}\right)^{-2},\hspace{1mm} i = 1, 2, \hdots, M
\end{equation}
with 
\begin{equation}
\Phi(\omega_0) = 2^k \cdot 10^{-6}
\end{equation}
and $k \in \{0, 2, 4, 6, 8\}$. We say that the road is of class $A$, if $k=0$, of class $B$, if $k=2$, up to class $E$ with $k=8$. It is obvious then that a higher value for $k$ results in a higher general amplitude of the road displacement $r(s)$ for all $s \in \mathbb{S}$ and therefore describes a road with a higher degree of roughness.  Furthermore the frequency domain is defined by the vector $\omega = (\omega_0, \omega_1,\hdots,\omega_M)^T$ with $\omega_0 = 1$, $\omega_1 = 0.02 \pi$, $\omega_M = 6\pi$ and $\omega_i = \omega_1 + (i-1) \cdot \Delta \omega$ for $i \in \{2,3,\hdots,M-1\}$, where $\Delta \omega = \frac{\omega_M-\omega_1}{M-1}$. The phase is given by realisations $\varphi_i$ with $i \in \{1,2,\hdots, M\}$ of a uniformly distributed random variable $\varphi \sim \mathcal{U}([0,2\pi))$. \\
In order to discuss time-dependent dynamical models, we need to switch from a constantly increasing distance over time described by set $\mathbb{S}$ to a time domain $[t_0, t_N]$. Therefore we assume that the vehicle drives with constant velocity $v > 0$. It is obvious that for all $s \in \mathbb{S}$, we have $s = t \cdot v$ for all $t \in [t_0, t_N]$.
And consequently, the road profile can time-dependently be defined by
\begin{equation}
r(t) = \sum_{i=1}^M A_i \sin(\omega_i t v - \varphi_i).
\end{equation}
The state can now be used to induce a force as described by Eq.$(1)$.

\subsection{Quarter-Car-Model}
In this section we introduce a system of second order ordinary differential equation that can be employed as a Quarter-Car-Model \cite{Faheem2006, Kulkarni2017}, which constitutes an approximation of a Half-Car-Model \cite{Abbas2013} or Full-Car-Model \cite{Mitra2013}. Our parameters are taken from \cite{Kumar2018}. The accelerations of the three rigid bodies of the QCM can then be described by the following equations 
\begin{align}
\ddot{z} &= -\frac{C_3}{m_3} \left( \dot{z}-\dot{y} \right) -\frac{K_3}{m_3} \left( z-y \right), \\
\ddot{y} &= -\frac{C_3}{m_2} \left( \dot{y}-\dot{z} \right) -\frac{C_2}{m_2}\left( \dot{y}-\dot{x} \right) -\frac{K_3}{m_2}\left( y-z \right)-\frac{K_2}{m_2}\left( y-x \right),\\
\ddot{x} &= -\frac{C_2}{m_1} \left( \dot{x}-\dot{y} \right) -\frac{K_2}{m_1}\left( x-y \right) -\frac{K_1}{m_1}\left( x-r \right),
\end{align}
where $C_2$ and $C_3$ are the damping constants with $C_2 = 4741 \frac{\textrm{Ns}}{\textrm{m}}$ and $C_3 =  615 \frac{\textrm{Ns}}{\textrm{m}}$, $K_1, K_2, K_3$ are the spring constants with $K_1 = 40000 \frac{\textrm{N}}{\textrm{m}}$, $K_2 = 149171 \frac{\textrm{N}}{\textrm{m}}$ and $K_3 = 98935 \frac{\textrm{N}}{\textrm{m}}$ and $m_1 = 145 \textrm{kg}$ is the mass of the wheel suspensions, $m_2 = 2160 \textrm{kg}$ the mass of the car body and $m_3 \in \{50, 51, \hdots, 200\} \textrm{kg}$ the mass of the passenger plus the seat's mass. A more complex dynamical human model can be developed following \cite{Abbas2010} but the simple version is sufficient for the observations considered in this approach. Furthermore, the states $x, y, z$ describe the relative displacement of the rigid bodies with masses $m_1$ for $x$, $m_2$ for $y$ and $m_3$ for $z$.\\
The velocities are given by $\dot{x} = \frac{\mathrm{d}x}{\mathrm{d}t}$, $\dot{y} = \frac{\mathrm{d}y}{\mathrm{d}t}$ and $\dot{z} = \frac{\mathrm{d}z}{\mathrm{d}t}$ and finally the accelerations by $\ddot{x} = \frac{\mathrm{d}^2x}{\mathrm{d}t^2}$, $\ddot{y} = \frac{\mathrm{d}^2y}{\mathrm{d}t^2}$ and $\ddot{z} = \frac{\mathrm{d}^2z}{\mathrm{d}t^2}$. The term $r$ describes the displacement of the road as defined in  Section $2.1$.\\
\\
\begin{figure}[!h]
\centering    	
\includegraphics[width=6.5cm]{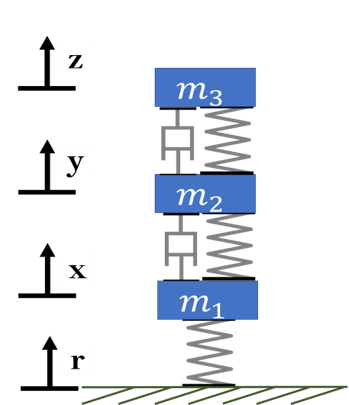}
\caption{Scheme of a Quarter-Car-Model: The approximative model of a car consists of the wheel suspensions (mass $m_1$ with state $x$), the car body (mass $m_2$ with state $y$) and the passenger on seat (mass $m_3$ with state $z$). The states are given by vertical displacement over time. The rigid masses are coupled using a total number of three springs and two dampers. A dynamical behaviour is experienced by the displacement of the road $r$ that results in a movement of the coupled mass-spring-damper system.}    
\end{figure}
In a less formal way, the vehicle model can be described as follows: A vehicle, as schematically given by Fig.1, drives along a specified road, whose displacement at a time $t \in [t_0, t_N]$ is given by a real valued scalar term $r(t)$. This displacement drives the dynamics of the wheel suspensions $x(t)$ and from there moves the remaining components $y(t)$ and $z(t)$ in the coupled system. The states are connected hierarchically by elements called springs and dampers. Regarding Eq.$(7)-(9)$, we see that the number of coefficients corresponds to the number of connected springs and dampers as shown in Fig.1. 
Due to the coupled structure of the model, the rigid bodies are not reacting simultaneously, but following a hierarchical, time-dependent structure.\\
Referring back to the initial statement that a dynamical system can be represented by Eq.$(1)$, it is obvious to see that the system being described by Eq.$(7)-(9)$ can be equivalently written in the form  
\begin{equation}
\underbrace{
\left(
\begin{matrix}
\ddot{z}\\
\ddot{y} \\
\ddot{x} \\
\dot{z} \\
\dot{y} \\
\dot{x}
\end{matrix}
\right)
}_{=: \dot{\rho}}
= 
\underbrace{ 
\left[
\begin{matrix}
-\frac{C_3}{m_3} & \frac{C_3}{m_3} & 0 & -\frac{K_3}{m_3} & \frac{K_3}{m_3} & 0 \\
\frac{C_3}{m_2} & -\frac{C_2+C_3}{m_2} & \frac{C_2}{m_2} & \frac{K_3}{m_2} & -\frac{K_2+K_3}{m_2} & \frac{K_2}{m_2} \\
0 & \frac{C_2}{m_1} & - \frac{C_2}{m_1} & 0 & \frac{K_2}{m_1} & - \frac{K_1 + K_2}{m_1} \\
1& 0 & 0 & 0 & 0 & 0 \\
0 & 1 & 0 & 0 & 0 & 0 \\
0 & 0 & 1 & 0 & 0 & 0 
\end{matrix}
\right]
}_{=: A}
\cdot
\underbrace{
\left(
\begin{matrix}
\dot{z}\\
\dot{y} \\
\dot{x} \\
z \\
y \\
x
\end{matrix}
\right)
}_{=: \rho}
+
\underbrace{
\left(
\begin{matrix}
0\\
0 \\
\frac{K_1}{m_1}r \\
0\\
0\\
0
\end{matrix}
\right).
}_{=: f}
\end{equation}
The above Eq.$(10)$ can therefore be seen as a system of first order ordinary differential equations. We assume that the states have a continuous dynamical behaviour, thus the system can (numerically) be solved, if the initial values $\rho(t_0) = (\dot{z}(t_0), \dot{y}(t_0), \dot{x}(t_0), z(t_0), y(t_0), x(t_0))^T = (\dot{z}_0, \dot{y}_0, \dot{x}_0, z_0, y_0, x_0)^T$ are known. Hold in mind that it does not make sense to solve the system in parallel, for instance using a numerical approximation given by a multi-dimensional variant of the forward Euler method. Therefore, we need to use a structure-preserving, symplectic integration scheme.

\subsection{Dataset}

To generate a synthetic dataset, the above ODE, defined by Eq.$(10)$, is simulated by a symplectic Euler scheme for geometric integration \cite{Hairer2006}. This guarantees that the interdependent relation between the coupled states is considered when computing the approximate solution of the differential equations. For simplification, we set $\dot{z} = w$, $\dot{y} = v$ and $\dot{x} = u$, which then results in an exact description of Eq.$(10)$ as a first order non-homogeneous system of ODE. We can then use the following iterative structure to come to an appropriate numerical solution of our system:
\begin{equation}
\begin{aligned}
u_{k+1} &= u_k + h \left( \frac{C_2}{m_1} v_k  - \frac{C_2}{m_1} u_k  + \frac{K_2}{m_1} y_k - \frac{K_1 + K_2}{m_1} x_k +\frac{K_1}{m_1}r_k \right) \\
v_{k+1} &= v_k + h \left( \frac{C_3}{m_2} w_k -\frac{C_2+C_3}{m_2} v_k +  \frac{C_2}{m_2} u_{k+1} + \frac{K_3}{m_2} z_k  -\frac{K_2+K_3}{m_2} y_k  + \frac{K_2}{m_2} x_k \right)  \\
w_{k+1} &= w_k + h \left( -\frac{C_3}{m_3} w_k + \frac{C_3}{m_3} v_{k+1} - \frac{K_3}{m_3} z_k + \frac{K_3}{m_3} y_k \right) \\
x_{k+1} &= x_k + h u_{k+1}  \\
y_{k+1} &= y_k + h v_{k+1}  \\
z_{k+1} &= z_k + h w_{k+1} 
\end{aligned}
\end{equation}
We use a discrete time scheme to compute the approximate solution with the above equations. Note that for instance $u_k = u(t_k)$ with $t_k = kh$, where $k \in \{0,1,\hdots,N\}$ with $N \in \mathbb{N}$ the number of discrete time steps and $h = \frac{t_N-t_0}{N}$ the step-width. The same scheme is analogously applied to $v(t_k), w(t_k), x(t_k), y(t_k), z(t_k)$ and also to the road displacement $r(t_k)$ for all $k \in \{0,1,\hdots, N\}$. The above iterative structure ensures a representation of the dynamical behaviour of a QCM.\\
We can now generate a synthetic dataset using the above described symplectic scheme and further assume that the dynamic system is in equilibrium at $t = t_0$. Consequently, we have initially $u_0 = v_0 = w_0 = x_0 = y_0 = z_0 = 0$, interpreted as relative states / velocities to the corresponding rigid masses. We further assume that the following experiment can be described by the dataset: The spring and damping parameters are equal for all samples of the dataset. This means that we always regard the same vehicle for all samples. The shape of the road's displacement is equal for a total number of $L_m \in \mathbb{N}$ samples. For each sample, the mass of the seat and the passenger is drawn from a uniformly distributed random variable with constraint that the mass has to be integer-valued.\\
We compute $L_r \in \mathbb{N}$ random shapes of the road displacement as described in Section 2.1 for discrete time steps $t_k$. Deviations from profile to profile are guaranteed by the randomly drawn phase for all $M \in \mathbb{N}$ sine waves for all time steps. In addition to that, the degree of roughness, characterized by $\Phi(\omega_0)$ is also randomly drawn with respect to one of the five acceptable classes $A-E$. Then for one of the road profiles $r^j$, $j\in \{1,2,\hdots,L_r\}$, we generate $m_3^{ij}$, $i \in \{1,2,\hdots, L_m \}$ and apply the scheme given by Eq.$(11)$. Thus we get $\hat{u}^{ij}, \hat{v}^{ij}, \hat{w}^{ij}, \hat{x}^{ij}, \hat{y}^{ij}, \hat{z}^{ij} \in \mathbb{R}^N$ as discrete solution of the non-homogeneous system with road profile $r^j$ and mass $m_3^{ij}$.\\
Then the discrete accelerations $\hat{\ddot{z}}^{ij}$ and $\hat{\ddot{y}}^{ij}$ can be computed using Eq.$(7)$ and Eq.$(8)$ by
\begin{align}
\hat{\ddot{z}}^{ij} &= -\frac{C_3}{m_3^{ij}} \left( \hat{w}^{ij}-\hat{v}^{ij} \right) -\frac{K_3}{m_3^{ij}} \left(\hat{z}^{ij}-\hat{y}^{ij} \right), \\
\hat{\ddot{y}}^{ij} &= -\frac{C_3}{m_2} \left( \hat{v}^{ij}-\hat{w}^{ij} \right) -\frac{C_2}{m_2}\left( \hat{v}^{ij}-\hat{u}^{ij} \right) -\frac{K_3}{m_2}\left( \hat{y}^{ij}-\hat{z}^{ij} \right)-\frac{K_2}{m_2}\left( \hat{y}^{ij}-\hat{x}^{ij} \right).
\end{align}
The discrete accelerations $\hat{\ddot{z}}^{ij}$ and $\hat{\ddot{y}}^{ij}$ can then be interpreted to be recordings of two g-Sensors, one measuring acceleration of the seat and one measuring acceleration of the car body in vertical direction. Therefore, computing the system's accelerations for different passenger's masses, simulates a car driving on different roads with different passengers.\\
We can then define our labelled dataset by
\begin{equation*}
\mathbb{X} = \left\{\left( \hat{\ddot{z}}^{ij}, \hat{\ddot{y}}^{ij}\right) \in \mathbb{R}^{N \times 2} \hspace{1mm} \vert \hspace{1mm} i \in \{1,2,\hdots,L_m\}, j \in \{1,2,\hdots,L_r\}\right\},
\end{equation*}
\begin{equation*}
\mathbb{M} = \left\{m^{ij} \in \{50,51,\hdots,200\} \hspace{1mm} \vert \hspace{1mm} i \in \{1,2,\hdots,L_m\}, j \in \{1,2,\hdots,L_r\}\right\},
\end{equation*}
both with a cardinality of $L = L_r \cdot L_m$ samples per set. Then the labelled dataset can be described by $\left( \mathbb{X}, \mathbb{M} \right) \subset \mathbb{R}^{L \times N \times 2} \times \mathbb{R}^L$. Hold in mind that $\left( \hat{\ddot{z}}^{ij}, \hat{\ddot{y}}^{ij}\right)$ for $i \in \{1,2,\hdots,L_r\}$ corresponds to the vertical acceleration that results in the Quarter-Car-Model by the road profile $r^j$ for $j \in \{1,2,\hdots, L_r\}$. Then we can separate the index set $\{1,2,\hdots, L_r\}$ that identifies the road profiles using an integer-valued separator $L_b \leq L_r$ such that we can define the labelled training dataset by
\begin{equation*}
\mathbb{X}_{train} = \left\{\left( \hat{\ddot{z}}^{ij}, \hat{\ddot{y}}^{ij}\right) \in \mathbb{R}^{N \times 2} \hspace{1mm} \vert \hspace{1mm} i \in \{1,2,\hdots,L_m\}, j \in \{1,2,\hdots,L_b\}\right\},
\end{equation*}
\begin{equation*}
\mathbb{M}_{train} = \left\{m^{ij} \in \{50,51,\hdots,200\} \hspace{1mm} \vert \hspace{1mm} i \in \{1,2,\hdots,L_m\}, j \in \{1,2,\hdots,L_b\}\right\},
\end{equation*}
and the labelled test dataset by
\begin{equation*}
\mathbb{X}_{test} = \left\{\left( \hat{\ddot{z}}^{ij}, \hat{\ddot{y}}^{ij}\right) \in \mathbb{R}^{N \times 2} \hspace{1mm} \vert \hspace{1mm} i \in \{1,2,\hdots,L_m\}, j \in \{L_b+1, L_b+2,\hdots,L_r\}\right\},
\end{equation*}
\begin{equation*}
\mathbb{M}_{test} = \left\{m^{ij} \in \{50,51,\hdots,200\} \hspace{1mm} \vert \hspace{1mm} i \in \{1,2,\hdots,L_m\}, j \in \{L_b+1, L_b+2,\hdots,L_r\}\right\}.
\end{equation*}
Now it can be guaranteed that there is no road profile used to generate training data simultaneously used for the test dataset. As far as our investigations are concerned, we choose the parameters $N=6000$, $L_r = 100$, $L_m = 100$ and $L_b = 8000$. Consequently the dataset simulates a Quarter-Car-Model on a total number of $100$ roads, where for each road there are $100$ passengers with arbitrary weight in the previously defined interval. A total number of $80$ roads are used for generating the training dataset and $20$ for the test dataset.\\
 
\subsection{Neural Network}
The previous section described in precise, how an appropriate sequential dataset that represents realizations of a coupled mass-spring-damper system can be computed using a symplectic integration technique like the semi-implicit Euler method. Convolutional neural networks have shown to achieve comparable results like recurrent neural networks \cite{Bai2018} for sequential data processing \cite{Dorffner1996}. In addition, convolutional networks can be easily used for a set of sequential data simultaneously, using multiple input channels. Therefore, we also consider a convolutional network architecture for the dataset we developed in Section $2.3$.\\
Comparable to image recognition tasks, the pair $\left( \hat{\ddot{z}}, \hat{\ddot{y}}\right)$ can numerically be processed like one row of an MNIST sample \cite{Chen2018}, using a one-dimensional convolutional operation. In this case, one dimensional means that there is only one direction the convolution is applied to.\\
We use augmentation strategies combined with batch-optimization to achieve a better generalization performance for unknown test samples. Therefore, instead of choosing the complete sample $\left( \hat{\ddot{z}}, \hat{\ddot{y}}\right) \in \mathbb{R}^{N \times 2}$, where the discrete values of the acceleration are by definition as described in Eq.$(11)$ restricted to the index set $I := \{0,1,\hdots,N-1\}$. Then, a subset $I_l \subset I$ can be defined using a smaller frame of $\bar{N} = 500$ discrete steps with randomly drawn starting point index $l \in \{0, 1, \hdots, 5500\}$, due to $5500$ is the maximum index number, such that a vector of size $500$ can be described within the index set $I$.\\
In detail, the subinterval is given by $I_l = \{l , l+1, \hdots, l+\bar{N}-1\}$. Therefore, following the notation of Section $2.3$, we can define the randomly chosen subset of the input by 
\begin{equation}
\left( \hat{\ddot{z}}, \hat{\ddot{y}}\right)_{I_l} =
\left(
\begin{matrix}
\hat{\ddot{z}}_l & \hat{\ddot{y}}_l \\
\hat{\ddot{z}}_{l+1} & \hat{\ddot{y}}_{l+1} \\
\vdots & \vdots \\
\hat{\ddot{z}}_{l+\bar{N}-1} & \hat{\ddot{y}}_{l+\bar{N}-1} 
\end{matrix}
\right).
\end{equation}
Assume that there are two unknown entries within the system matrix $A$, given by the two-dimensional vector
\begin{equation}
p = 
\left(
\begin{matrix}
p_1 \\
p_2
\end{matrix}
\right)
= 
\left(
\begin{matrix}
\frac{C_3}{m_3} \\
\frac{K_3}{m_3}
\end{matrix}
\right),
\end{equation}
that consists of the two parameters to describe the acceleration of the passenger $\ddot{z}$ in Eq.$(10)$. \\
Then a deep convolutional neural network can be defined by the following function
\begin{align}
g_\theta: \mathbb{R}^{500 \times 2}  &\longrightarrow \mathbb{R}^2 \\
\left(\hat{\ddot{z}}, \hat{\ddot{y}} \right)_{I_{\cdot}} & \longmapsto \left( \begin{matrix} \tilde{p}_1 \\ \tilde{p}_2 \end{matrix} \right),
\end{align}
where $\theta$ describes the set of weights and biases of the convolutional neural network, $\tilde{p}_1$ is the prediction of the first parameter $p_1$ and analogously $\tilde{p}_2$ is the prediction for the second parameter $p_2$. This means that there are two unknown parameters within our system matrix $A$, we want the neural network to predict from the dataset. \\
Therefore, we use the following network structure: The first layer is a convolutional one with input dimension according to Eq.$(16)$ with $100$ filters of size $50$ and the hyperbolic tangent as activation function, followed by a second convolutional layer with again $100$ filters but with size $10$.  The output is then flattened and further processed using three fully-connected layers with $100$, $10$ and $2$ neurons with fitting weighting matrices and biases. For the output layer, we map the computational results of weighting and biasing to the absolute value in order to guarantee positive predictions of the parameters.

\section{Numerical Examples}
Following the notation of the previous section, the output of the neural network can be denoted by $g_\theta(\hat{\ddot{z}}, \hat{\ddot{y}}) = \left(\tilde{p}_1, \tilde{p}_2 \right)^T = \tilde{p}$. Therefore, the neural network's output can be separated into the prediction for the first parameter $\tilde{p}_1 = g_{\theta;1}(\hat{\ddot{z}}, \hat{\ddot{y}})$ and analogously for the second one $\tilde{p}_2=g_{\theta;2}(\hat{\ddot{z}}, \hat{\ddot{y}})$ Then the neural network to predict these two parameters can be trained using the labelled objective
\begin{equation}
J_L(\tilde{p}, p, \theta) = \sum_{i=1}^d |p_i-g_{\theta;i}(\hat{\ddot{z}}, \hat{\ddot{y}})|,
\end{equation}
where for this special case we have $d=2$ being the number of parameters to predict.\\
Furthermore, we know by Eq.$(7)$ that the acceleration of the passenger can be computed by
\begin{align}
\ddot{z} &= -\frac{C_3}{m_3}(\dot{z}-\dot{y}) - \frac{K_3}{m_3} (z-y) \\
&= -p_1 (\dot{z}-\dot{y}) - p_2 (z-y).
\end{align}
We assume that the device that records the acceleration data delivers the acceleration without noise, or more realistic, we assume that the acceleration data already has been de-noised. Then it is possible to get a discrete approximation of the velocities $\dot{y}, \dot{z}$ and consequently also approximations of the states $y, z$ with a numerical integration scheme. For a discrete time frame $t_0 < t_1 < \hdots < t_N$ with $h = t_{k+1}-t_k$ for all $k \in \{0,1,\hdots,N-1\}$ the recorded acceleration is given by
\begin{equation}
\hat{\ddot{z}}_k = \ddot{z}(t_k)
\end{equation}
and the same holds for the recorded acceleration of the car body $\hat{\ddot{y}}_k$. Consequently, using a symplectic Euler integration scheme, we get $\hat{\dot{z}}_k$ for all $k \in \{1,2,\hdots, N-1\}$ by
\begin{align*}
\hat{\dot{z}}_k &= \hat{\dot{z}}_{k-1} +  \int_{t_{k-1}}^{t_{k}} \hat{\ddot{z}}_{k-1} \mathrm{d}s\\
&= \hat{\dot{z}}_{k-1} +  h \cdot \hat{\ddot{z}}_{k-1}
\end{align*}
and analogously the state $\hat{z}_k$ by
\begin{align*}
\hat{z}_k &= \hat{z}_{k-1} +  \int_{t_{k-1}}^{t_{k}} \hat{\dot{z}}_{k} \mathrm{d}s\\
&= \hat{z}_{k-1} +  h \cdot \hat{\dot{z}}_{k},
\end{align*}
which reduces the approximation error that follows from simple forward Euler integration \cite{Yang2006}.
Same scheme can be applied to compute the integrated values of the car body's acceleration $\hat{\ddot{y}}_k$. Then, using the output of the neural network for these specific values, we can make a prediction of the initial acceleration, which corresponds to the computation of the discrete acceleration for the dataset in Eq.$(12)$, by 
\begin{equation}
\tilde{\ddot{z}}_k = -g_{\theta;1}(\hat{\ddot{z}}, \hat{\ddot{y}})(\hat{\dot{z}}_k-\hat{\dot{y}}_k) - g_{\theta;2}(\hat{\ddot{z}}, \hat{\ddot{y}}) (\hat{z}_k-\hat{y}_k)
\end{equation}
for all $k \in \{0,1,\hdots,N-1\}$. We then get an objective for unlabelled learning by
\begin{equation}
J_U(\tilde{\ddot{z}}, \hat{\ddot{z}}, \theta) = \sum_{l \in I_l} |\hat{\ddot{z}}_{l} - (-g_{\theta;1}(\hat{\ddot{z}}, \hat{\ddot{y}})(\hat{\dot{z}}_{l}-\hat{\dot{y}}_{l}) - g_{\theta;2}(\hat{\ddot{z}}, \hat{\ddot{y}}) (\hat{z}_{l}-\hat{y}_{l}))|^2,
\end{equation} 
where $I_l$ is the randomly drawn index set described in Section $2.4$. Comparable to an auto-encoder neural network, the original data is rebuilt out of a data representation in a low-dimensional (latent) space \cite{Kingma2019}, here being described by the parameter vector $\tilde{p}$.\\
\subsection{Labelled System Identification}
In a first experiment, the objective to minimize is $J_L(\tilde{p}, p, \theta)$, Therefore we discuss the initial results of the training dataset. We statistically evaluate our methods, with respect to absolute and relative mean error with corresponding variance.\\
\begin{figure}[!h]
	\centering
    \begin{minipage}[t]{7.5cm}
    	\centering
    	\includegraphics[width=7.5cm]{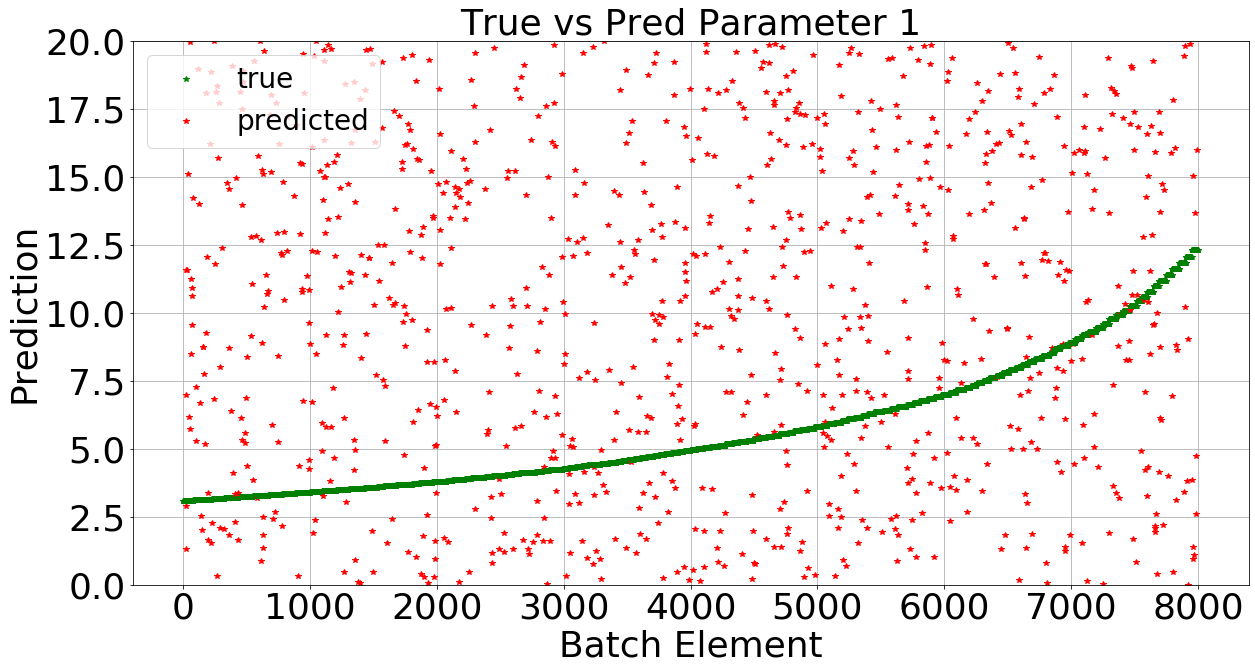}
    	\caption{Predictions (red) compared to true parameter values (green) for $p_1$ at initialization $i=0$ for the training dataset with the objective $J_L$.}
    \end{minipage}
    \hspace{1cm}
    \begin{minipage}[t]{7.5cm}
    	\centering
    	\includegraphics[width=7.5cm]{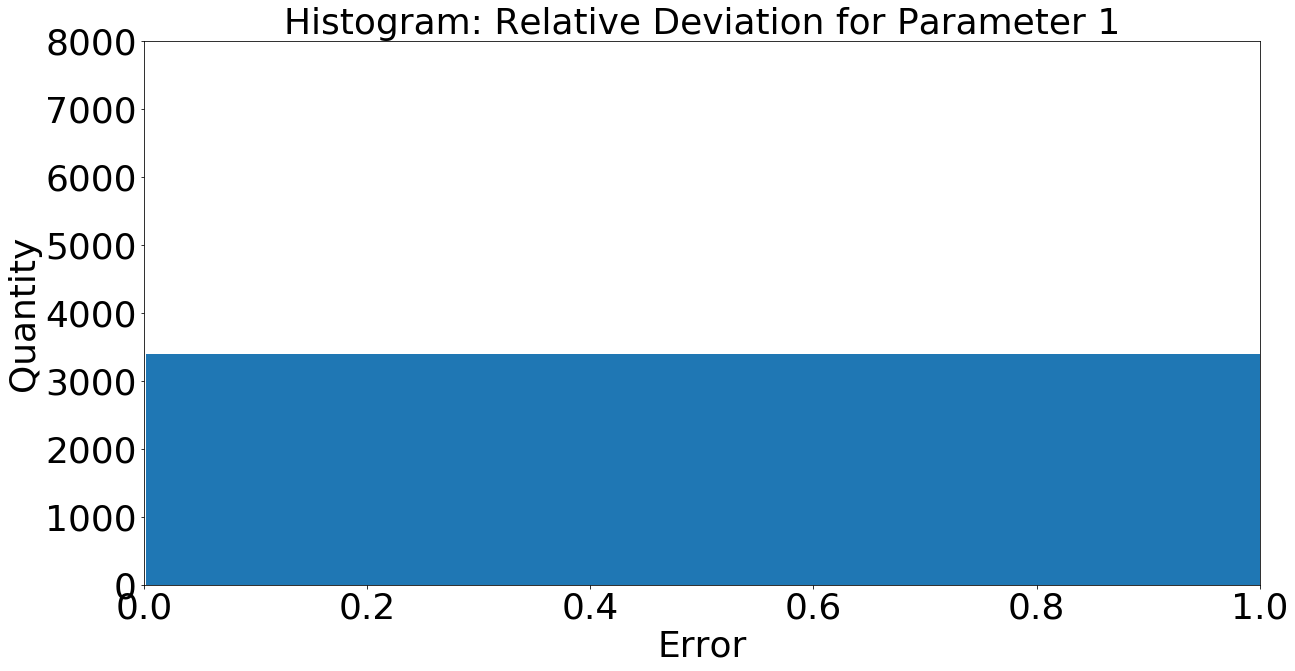}
    	\caption{Relative deviation of the prediction to the true parameter value for $p_1$ at initialization $i=0$ for the training dataset with the objective $J_L$.}
    \end{minipage}
     
    \begin{minipage}[t]{7.5cm}
    	\centering
    	\includegraphics[width=7.5cm]{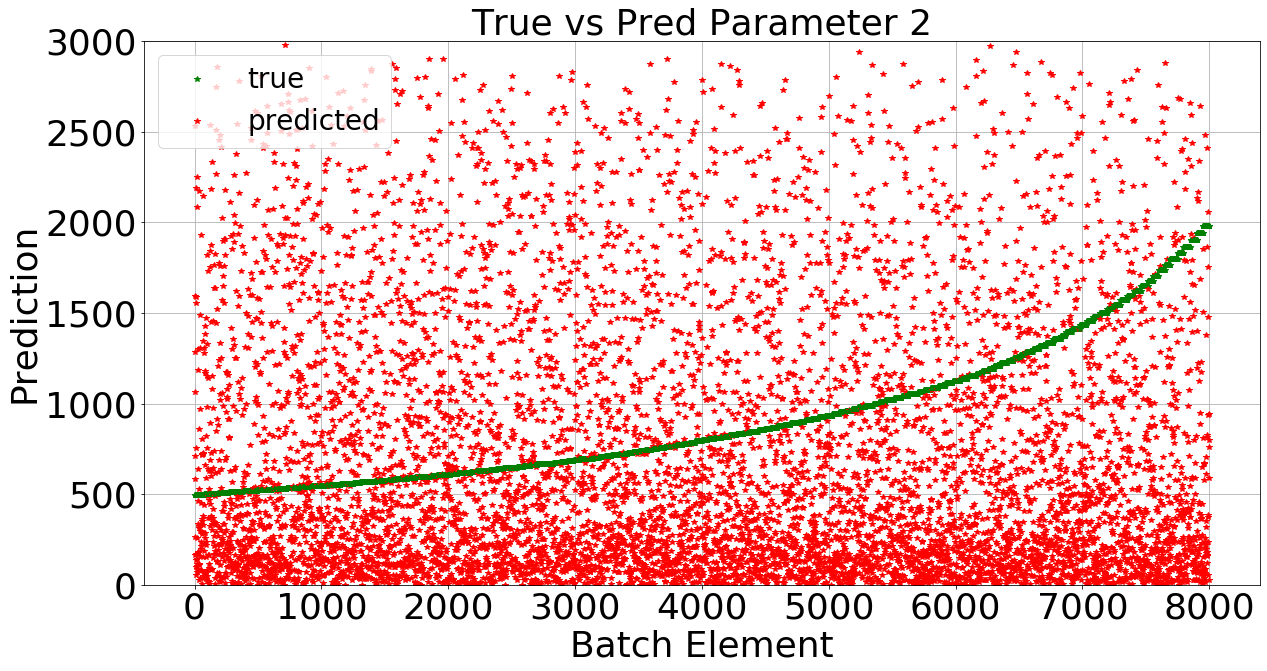}
    	\caption{Predictions (red) compared to true parameter values (green) for $p_2$ at initialization $i=0$ for the training dataset with the objective $J_L$.}
    \end{minipage}
    \hspace{1cm}
    \begin{minipage}[t]{7.5cm}
    	\centering
    	\includegraphics[width=7.5cm]{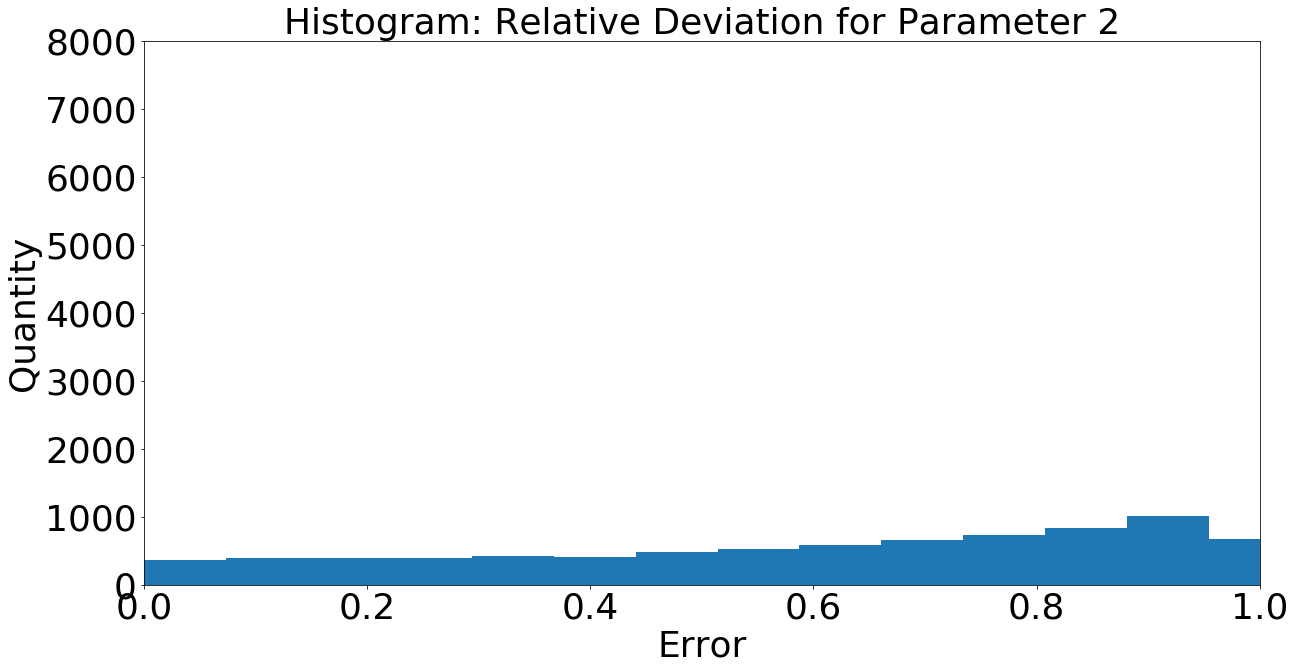}
    	\caption{Relative deviation of the prediction to the true parameter value for $p_2$ at initialization $i=0$ for the training dataset with the objective $J_L$.}
    \end{minipage}
\end{figure}
We predict the values of the parameters $p_1$ and $p_2$ for the training dataset with the help of a one-dimensional convolutional neural network, with objective $J_L$ for training. Fig.2 and Fig.4 show the predictions and the true values for $p_1$ (Fig.2) and for $p_2$ (Fig.4), where Fig.3 and Fig.5 show the relative deviations between the true parameter values and the predictions coming from the output layer of the neural network. More precisely, as far as the left-hand side is concerned, we sorted the training dataset according to the value of the true parameter. Therefore, we see numbers reaching from $1$ to $8000$ on the abscissa and a sequence of growing parameter values on the ordinate, represented by the green curve. This curve then describes the true parameter value for $p_1$ (Fig.2) and $p_2$ (Fig.4) for the entire training dataset. In contrast, the red points correspond to the network's prediction. Consequently, for each green point on the "curve" there is exactly one corresponding red point on the same vertical level. \\
Optimally, the red points should therefore fit the green line of the true parameter values. This property can be interpreted as prediction being equal to the underlying label of the input data.\\
Besides, the right-hand side shows the relative deviation of prediction to true parameter value plotted as histograms. The relative deviation of a true parameter $p_j$ with respect to the neural network's prediction $\tilde{p}_j$ for $j \in \{1,2\}$ is then given by 
\begin{equation}
\frac{\vert p_j - \tilde{p}_j \vert}{p_j}.
\end{equation}
The histograms then show the relative deviation on the abscissa, where the number of samples, that approximately lie within the same error range is shown on the ordinate. The four plots show the results at $i=0$ optimization steps. Therefore, we actually cannot see any valuable results, but the initialization is adequate to get an impression, how the optimal solution should look like. Weights and biases are randomly initialized, therefore we also get random outputs of the neural network. Optimally, the figures on the l.h.s. should show an approximation of the red point cloud to the green label line, where the histograms on the r.h.s. should concentrate close to zero on the x-axis, meaning we should have a small error for all samples of the training set.\\
\begin{figure}[!h]
	\centering
    \begin{minipage}[t]{7.5cm}
    	\centering
    	\includegraphics[width=7.5cm]{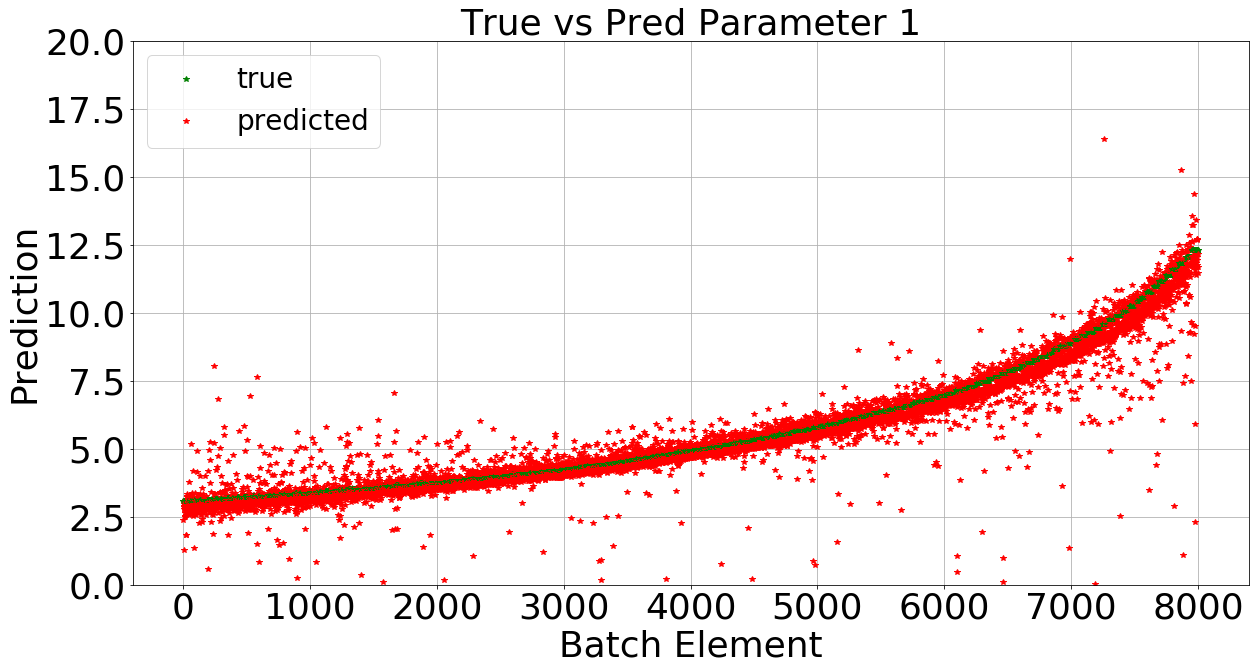}
    	\caption{Predictions (red) compared to true parameter values (green) for $p_1$ at optimization step $i=500.000$ for the training dataset with the objective $J_L$.}
    \end{minipage}
    \hspace{1cm}
    \begin{minipage}[t]{7.5cm}
    	\centering
    	\includegraphics[width=7.5cm]{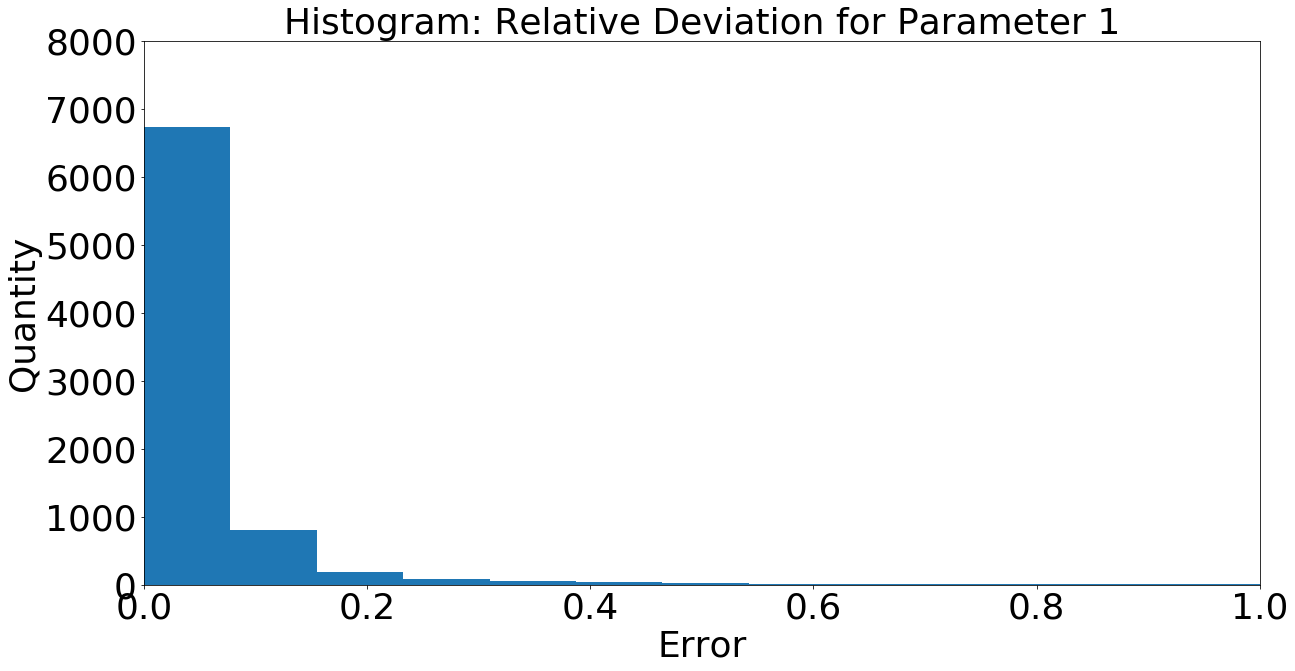}
    	\caption{Relative deviation of the prediction to the true parameter value for $p_1$ at optimization step $i=500.000$ for the training dataset with the objective $J_L$.}
    \end{minipage}
     
    \begin{minipage}[t]{7.5cm}
    	\centering
    	\includegraphics[width=7.5cm]{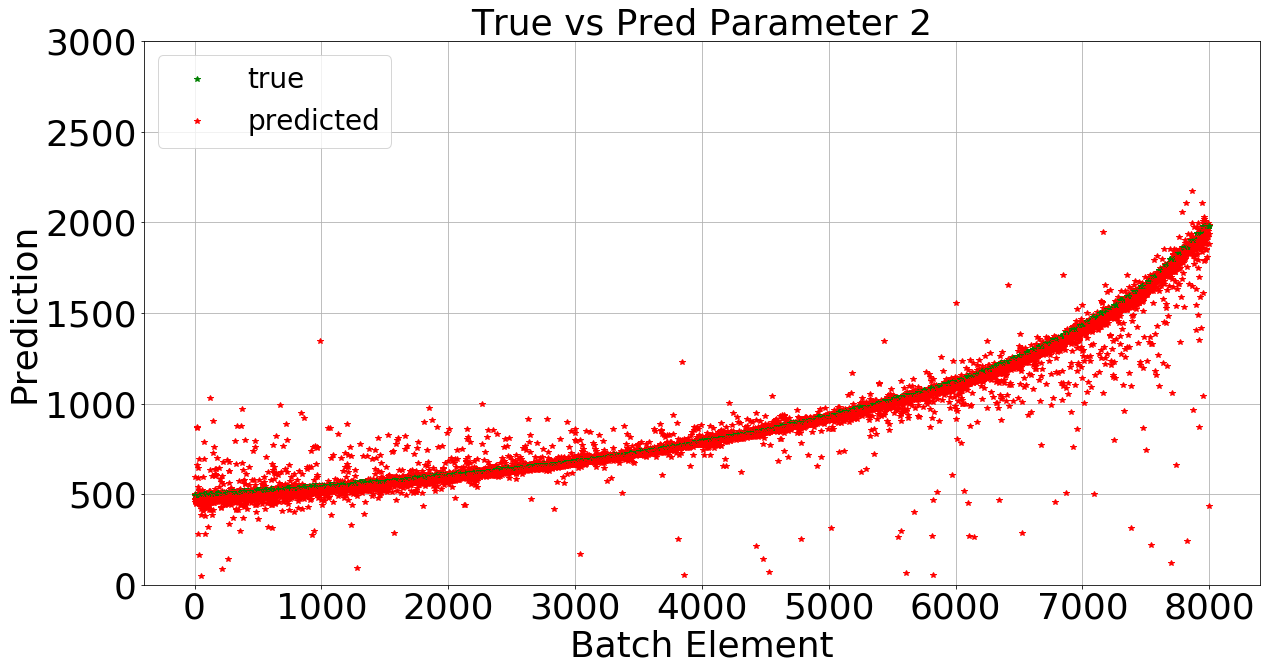}
    	\caption{Predictions (red) compared to true parameter values (green) for $p_2$ at optimization step $i=500.000$ for the training dataset with the objective $J_L$.}
    \end{minipage}
    \hspace{1cm}
    \begin{minipage}[t]{7.5cm}
    	\centering
    	\includegraphics[width=7.5cm]{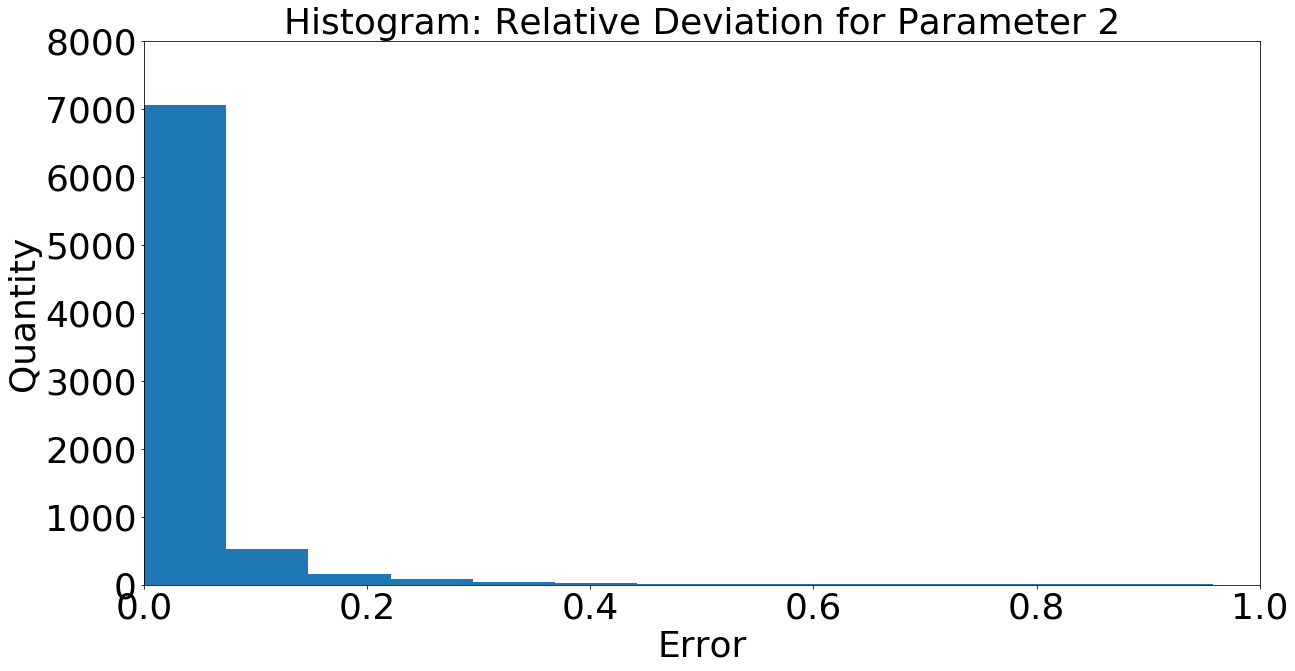}
    	\caption{Relative deviation of the prediction to the true parameter value for $p_2$ at optimization step $i=500.000$ for the training dataset with the objective $J_L$.}
    \end{minipage}
\end{figure}
The results in Fig.6 and Fig.7 show the prediction of the parameters and the true parameter values for the entire training dataset after $i = 500.000$ optimization steps, using Adam optimizer with learning-rate $\eta = 0.001$ and batch-size $M = 100$. As already described for $i = 0$ optimization steps, while training, the red point cloud should converge to the green label line. As we can see for both Fig.6 as well as for Fig.8, the red cloud indeed comes closer to the optimal label values.\\
In addition to that, also the histograms of the relative deviation, as can be seen in Fig.7 and Fig.9, show the expected result: The main samples have a relative deviation close to zero as can be seen by a left-skewed distribution of the relative deviation. Precisely, we have a relative mean deviation of $\mu = 0.062$ and a mean standard deviation of this error of $\sigma = 0.091$. For the second parameter, we have $\mu = 0.043$ and $\sigma = 0.083$.\\
The architecture of the convolutional neural network is therefore sufficient as far as the task of labelled learning for the given dataset is concerned. \\
\begin{figure}[!h]
	\centering
    \begin{minipage}[t]{7.5cm}
    	\centering
    	\includegraphics[width=7.5cm]{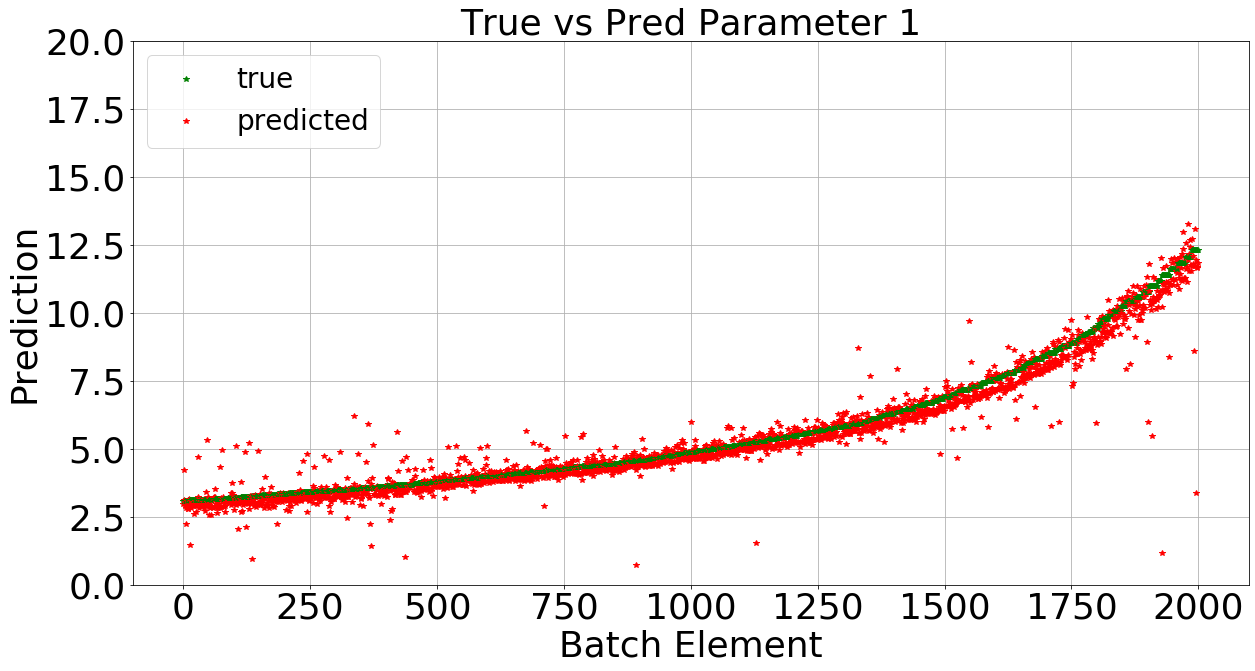}
    	\caption{Predictions (red) compared to true parameter values (green) for $p_1$ at optimization step $i=500.000$ for the test dataset with the objective $J_L$.}
    \end{minipage}
    \hspace{1cm}
    \begin{minipage}[t]{7.5cm}
    	\centering
    	\includegraphics[width=7.5cm]{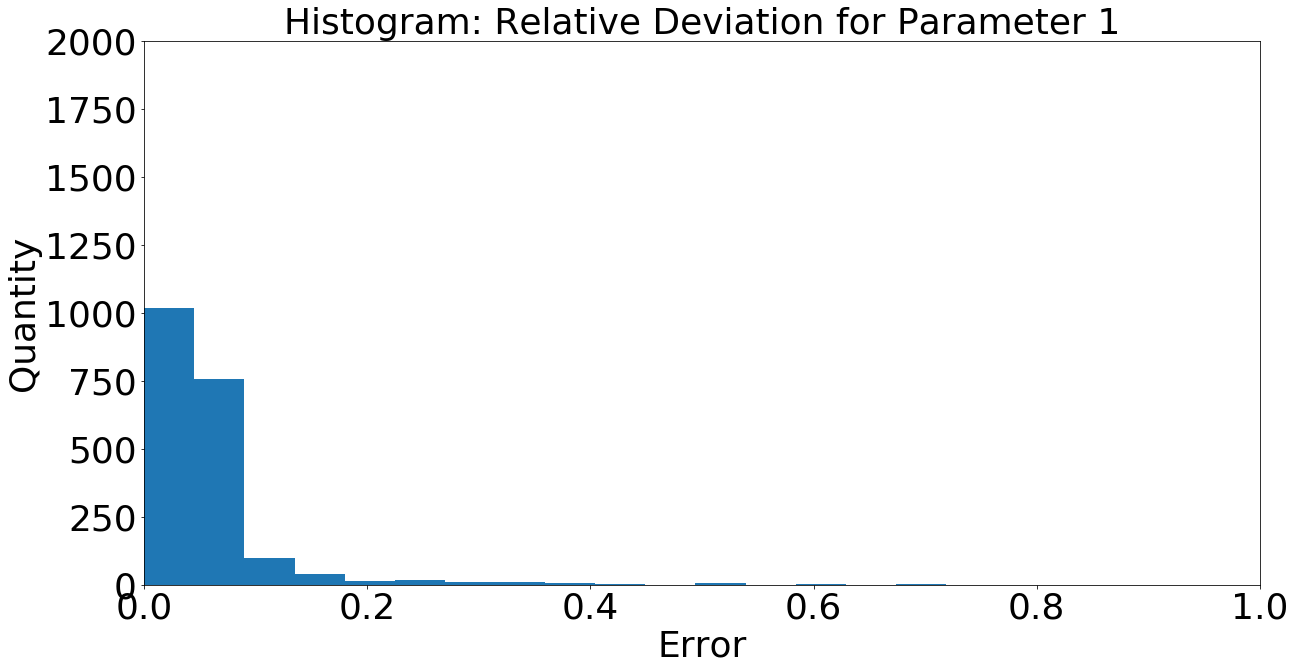}
    	\caption{Relative deviation of the prediction to the true parameter value for $p_1$ at optimization step $i=500.000$ for the test dataset with the objective $J_L$.}
    \end{minipage}
     
    \begin{minipage}[t]{7.5cm}
    	\centering
    	\includegraphics[width=7.5cm]{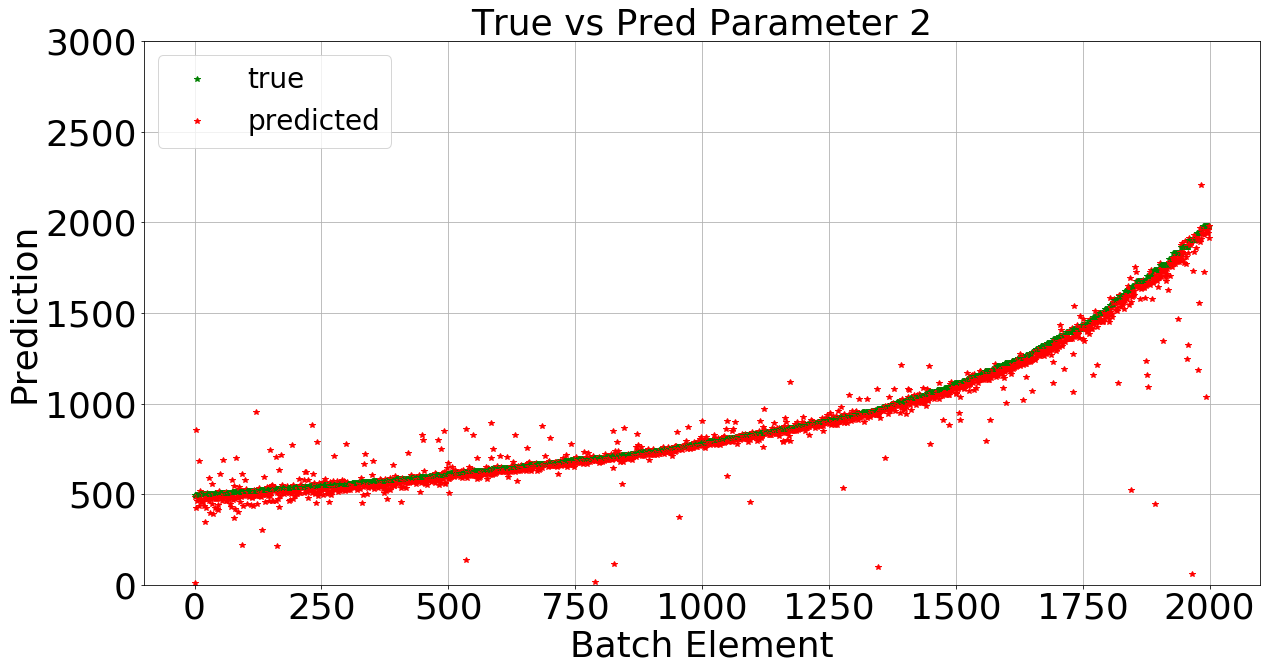}
    	\caption{Predictions (red) compared to true parameter values (green) for $p_2$ at optimization step $i=500.000$ for the test dataset with the objective $J_L$.}
    \end{minipage}
    \hspace{1cm}
    \begin{minipage}[t]{7.5cm}
    	\centering
    	\includegraphics[width=7.5cm]{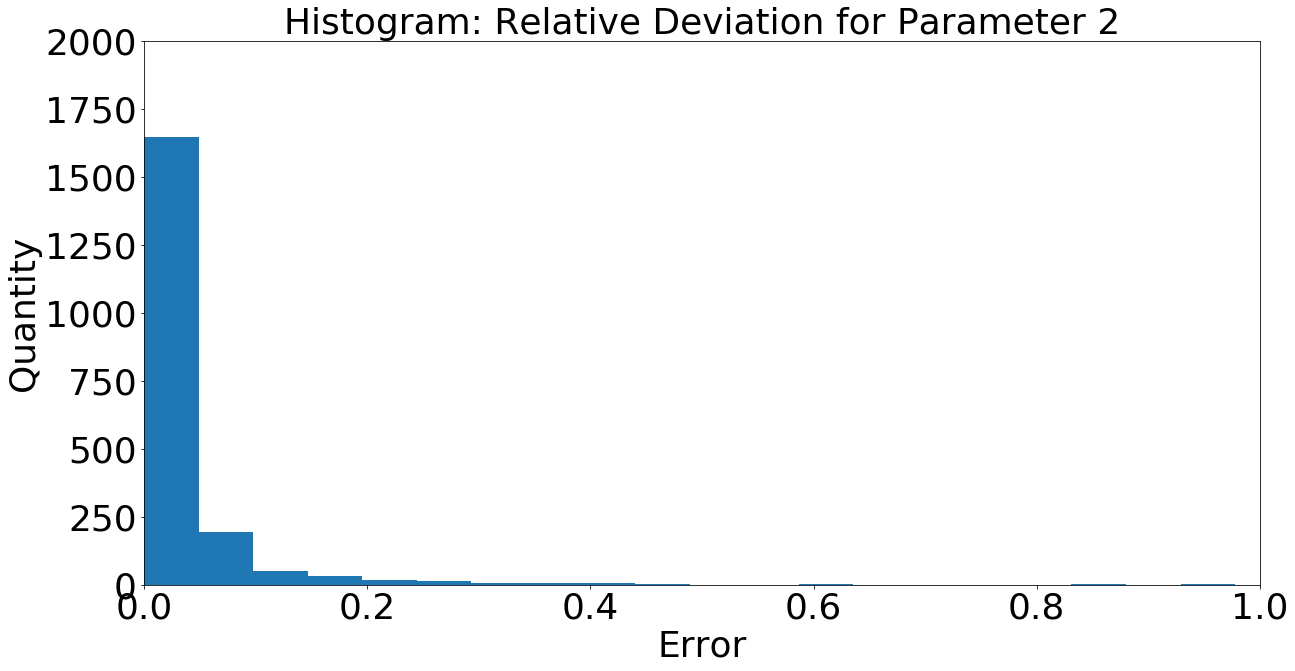}
    	\caption{Relative deviation of the prediction to the true parameter value for $p_2$ at optimization step $i=500.000$ for the test dataset with the objective $J_L$.}
    \end{minipage}
\end{figure}
Furthermore, we also want to discuss the generalization quality of the neural network. Therefore, we analyse the same plots as already discussed for the training data, but now for the unknown test dataset. Consequently, instead of $8000$ samples, as shown on the abscissa for Fig.2, Fig.4, Fig.6 and Fig.8, there is a range of $2000$ samples for the test data, as can be seen in Fig.10 and Fig.12. The results are similar to those of the training dataset. For the first parameter $p_1$ we have on average $\mu = 0.060$ and $\sigma = 0.079$, where for $p_2$ we get $\mu = 0.042$ and $\sigma = 0.084$.\\
In summary, we conclude that the mean training and test error are close to each other, as well as the corresponding standard deviation. Therefore, the convolutional neural network generalizes well with the given architecture using objective function $J_L$ for the data of our Quarter-Car-Model.

\subsection{Unlabelled System Identification}
Let us now assume that the true values of the parameters $p_1$ and $p_2$ are not known for any sample of the training dataset. Nevertheless, we want to identify the unknown parameters via usage of our deep convolutional neural network. Therefore, as initially described, we use our unlabelled objective function $J_U(\tilde{\ddot{z}}, \ddot{z}, \theta)$ to minimize the squared difference of $\hat{\ddot{z}}_l$ and the reproduction $\tilde{\ddot{z}}_l$ for $l \in I_l$.\\
\begin{figure}[!h]
	\centering
    \begin{minipage}[t]{7.5cm}
    	\centering
    	\includegraphics[width=7.5cm]{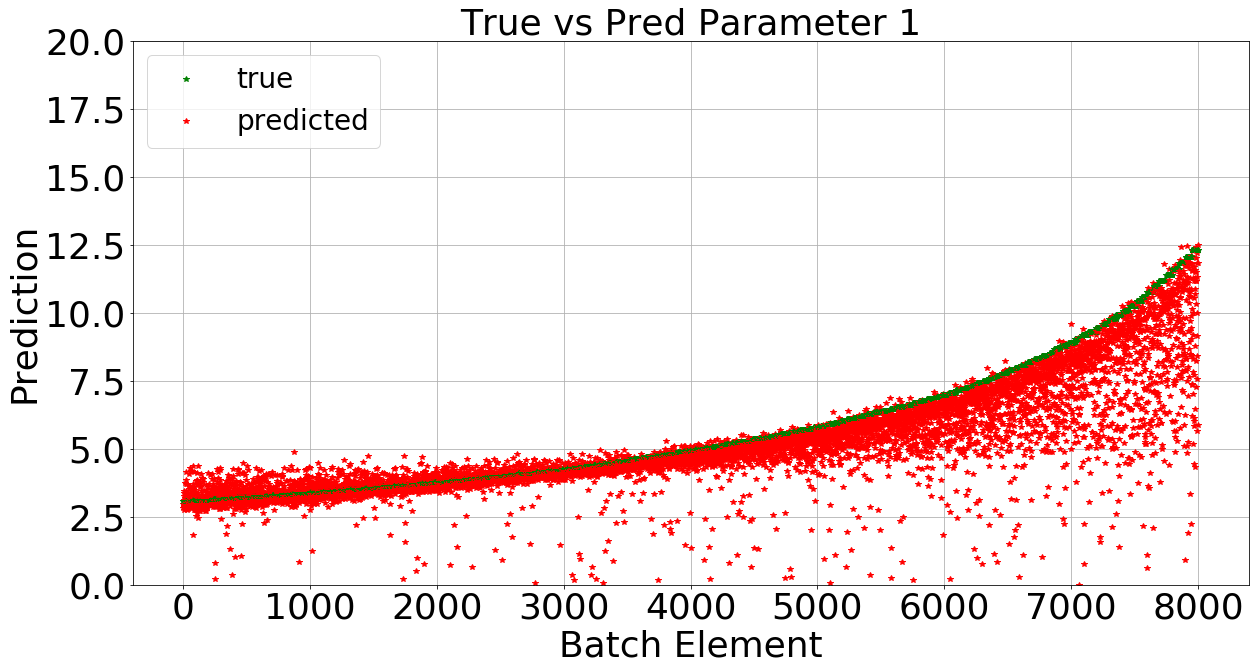}
    	\caption{Predictions (red) compared to true parameter values (green) for $p_1$ at optimization step $i=500.000$ for the training dataset with the objective $J_U$.}
    \end{minipage}
    \hspace{1cm}
    \begin{minipage}[t]{7.5cm}
    	\centering
    	\includegraphics[width=7.5cm]{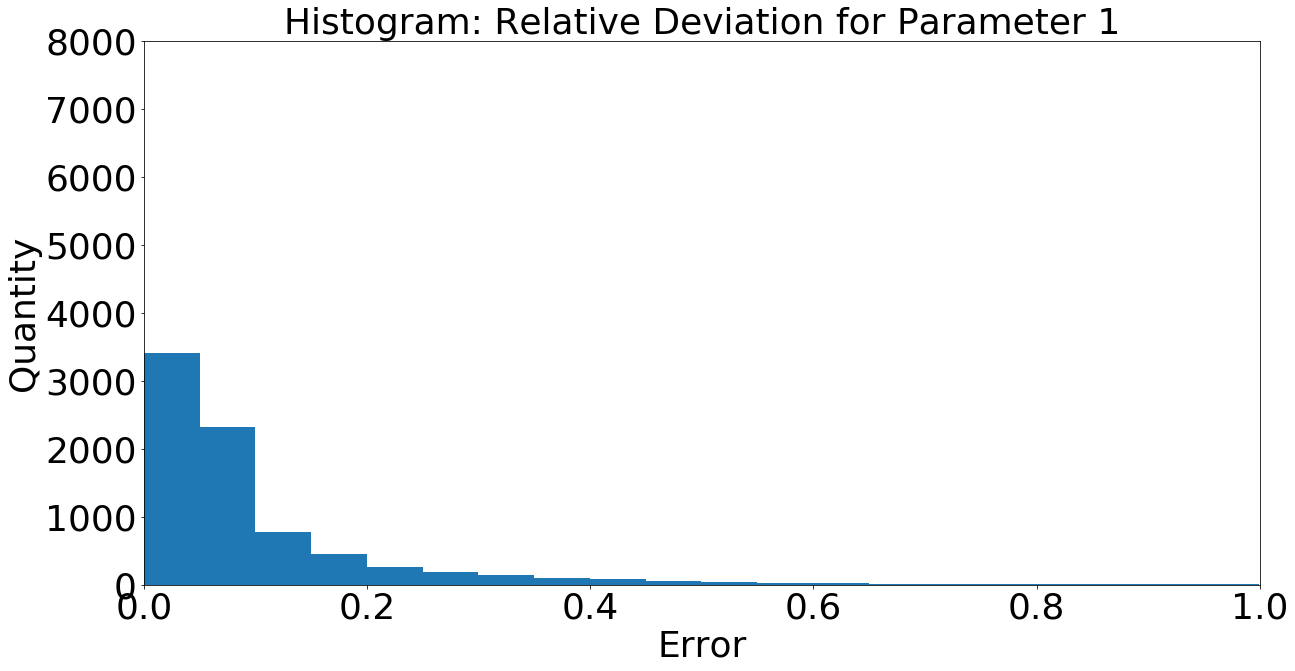}
    	\caption{Relative deviation of the prediction to the true parameter value for $p_1$ at optimization step $i=500.000$ for the training dataset with the objective $J_U$.}
    \end{minipage}
     
    \begin{minipage}[t]{7.5cm}
    	\centering
    	\includegraphics[width=7.5cm]{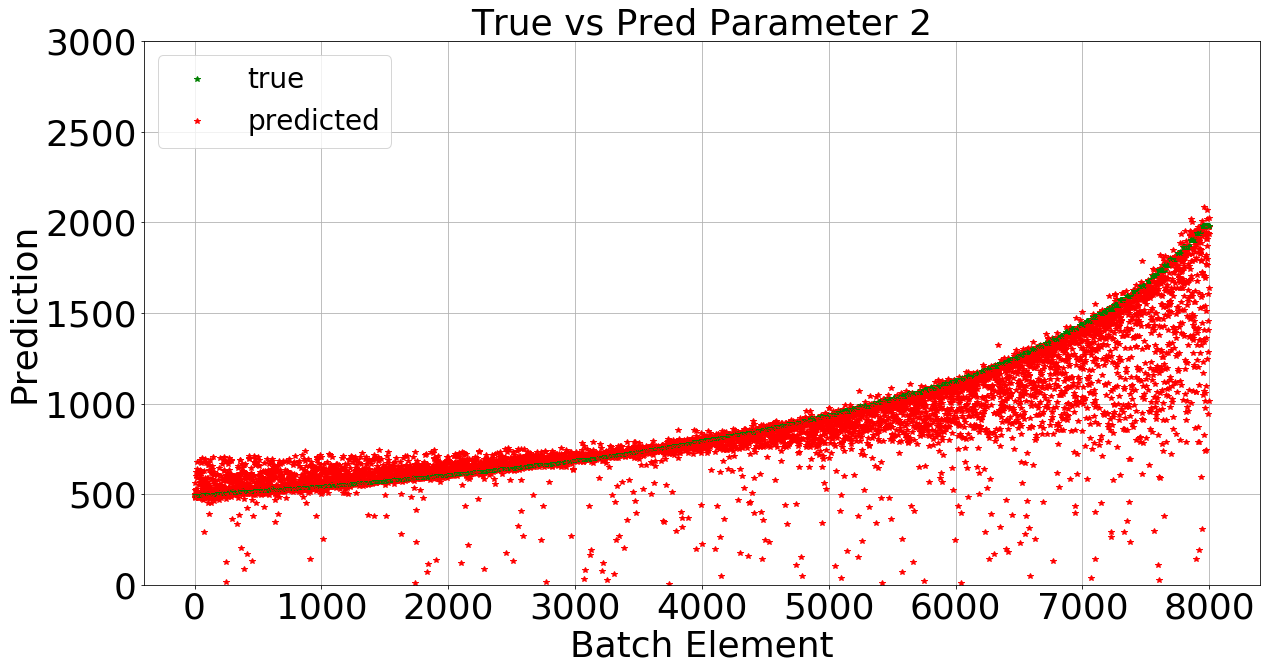}
    	\caption{Predictions (red) compared to true parameter values (green) for $p_2$ at optimization step $i=500.000$ for the training dataset with the objective $J_U$.}
    \end{minipage}
    \hspace{1cm}
    \begin{minipage}[t]{7.5cm}
    	\centering
    	\includegraphics[width=7.5cm]{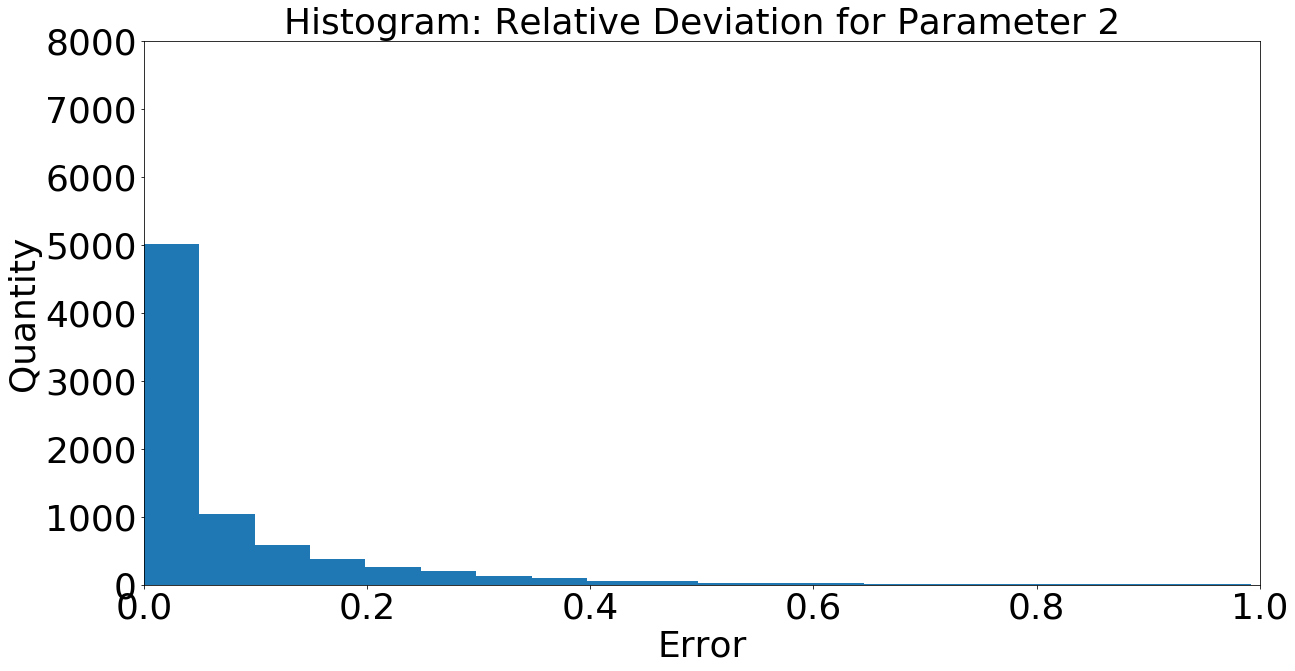}
    	\caption{Relative deviation of the prediction to the true parameter value for $p_2$ at optimization step $i=500.000$ for the training dataset with the objective $J_U$.}
    \end{minipage}
\end{figure}
Compared to the labelled approach, we skip the initialization analysis. We directly have a look at the usual plots for our experiments for the entire training data after $i = 500.000$ optimization steps. Again we use Adam optimization with the same learning-rate and batch-size as described for the first experiment. Here, the absolute deviation of parameter $p_1$ and $p_2$ are shown in Fig.$(14)$ and Fig.$(16)$, where the corresponding relative deviations are shown by the histograms in Fig.$(15)$ and Fig.$(17)$. Comparable to the first experiment, we see that the red points come closer to the green optimal label line, although we can recognize that there is a larger deviation, when regarding large parameter values for both $p_1$ as well as $p_2$. Again, the precise values are given by $\mu = 0.102$ and $\sigma = 0.131$ for $p_1$ and $\mu = 0.082$ and $\sigma = 0.131$ for $p_2$. Consequently, the mean error for the training data is approximately $4\%$ higher compared to the results of Section  $3.1$.\\
\begin{figure}[!h]
	\centering
    \begin{minipage}[t]{7.5cm}
    	\centering
    	\includegraphics[width=7.5cm]{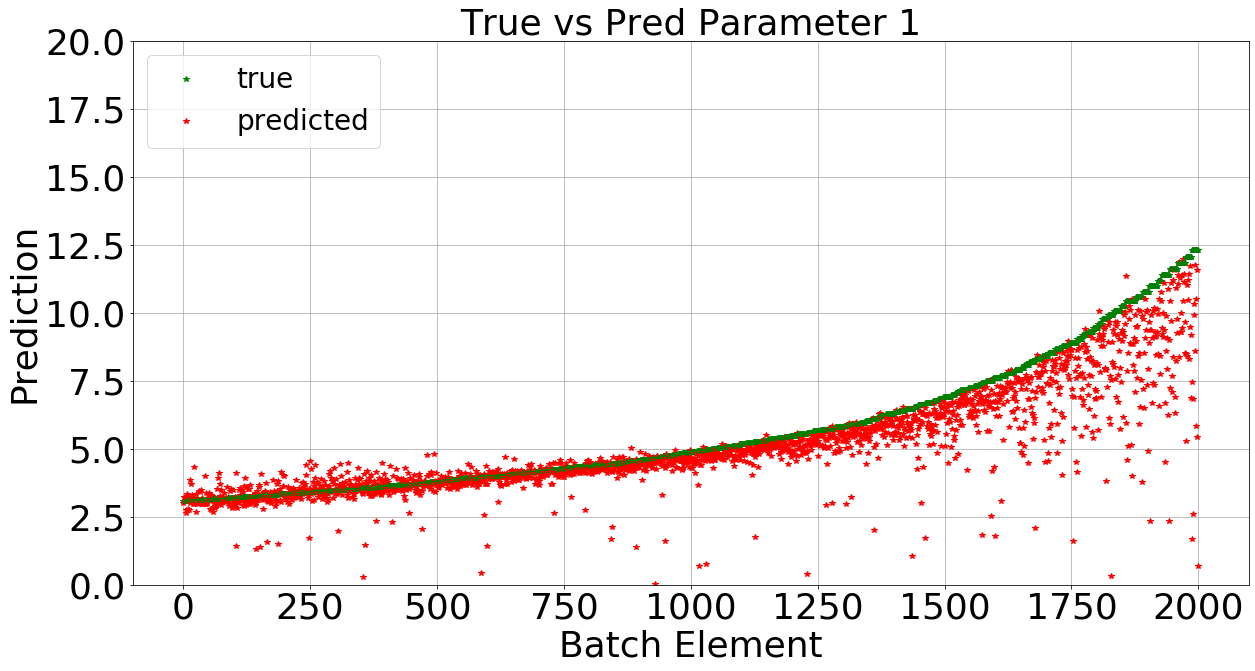}
    	\caption{Predictions (red) compared to true parameter values (green) for $p_1$ at optimization step $i=500.000$ for the test dataset with the objective $J_U$.}
    \end{minipage}
    \hspace{1cm}
    \begin{minipage}[t]{7.5cm}
    	\centering
    	\includegraphics[width=7.5cm]{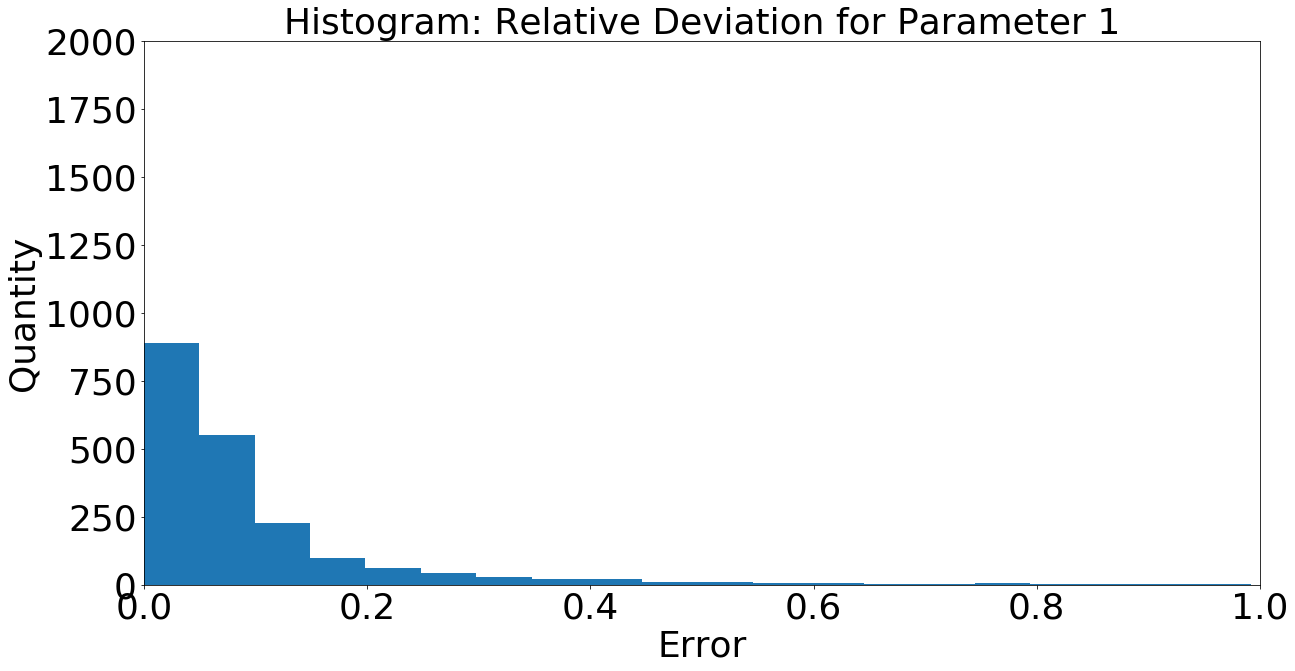}
    	\caption{Relative deviation of the prediction to the true parameter value for $p_1$ at optimization step $i=500.000$ for the test dataset with the objective $J_U$.}
    \end{minipage}
     
    \begin{minipage}[t]{7.5cm}
    	\centering
    	\includegraphics[width=7.5cm]{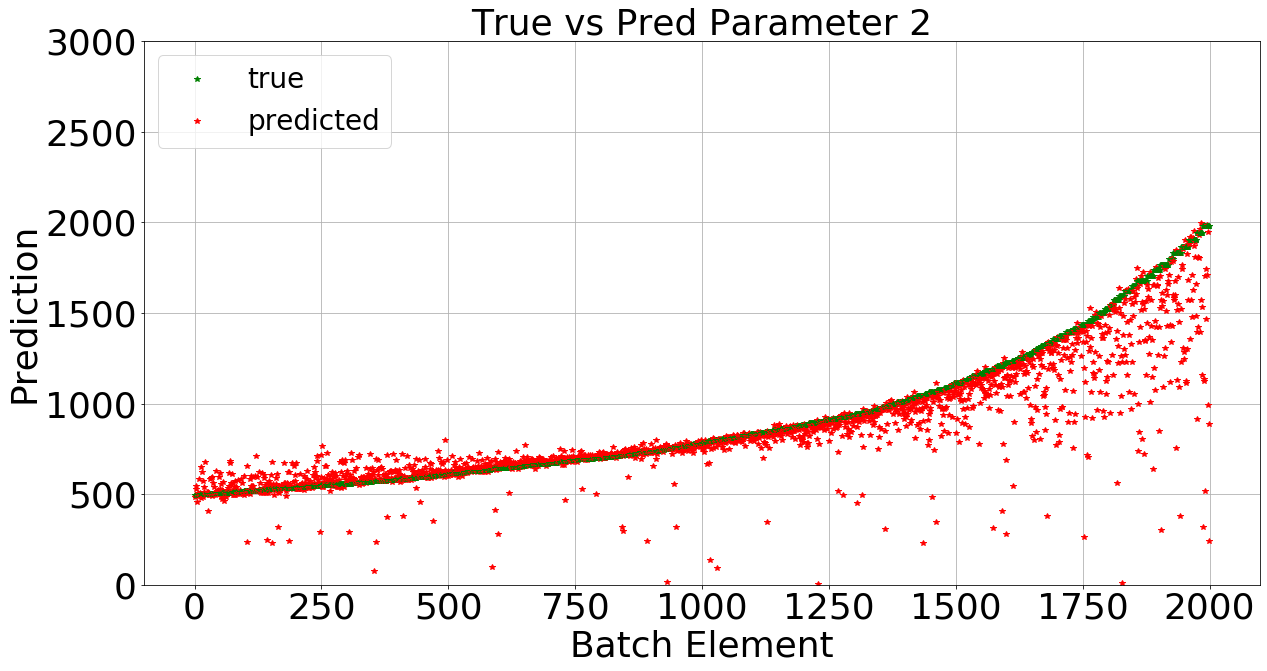}
    	\caption{Predictions (red) compared to true parameter values (green) for $p_2$ at optimization step $i=500.000$ for the test dataset with the objective $J_U$.}
    \end{minipage}
    \hspace{1cm}
    \begin{minipage}[t]{7.5cm}
    	\centering
    	\includegraphics[width=7.5cm]{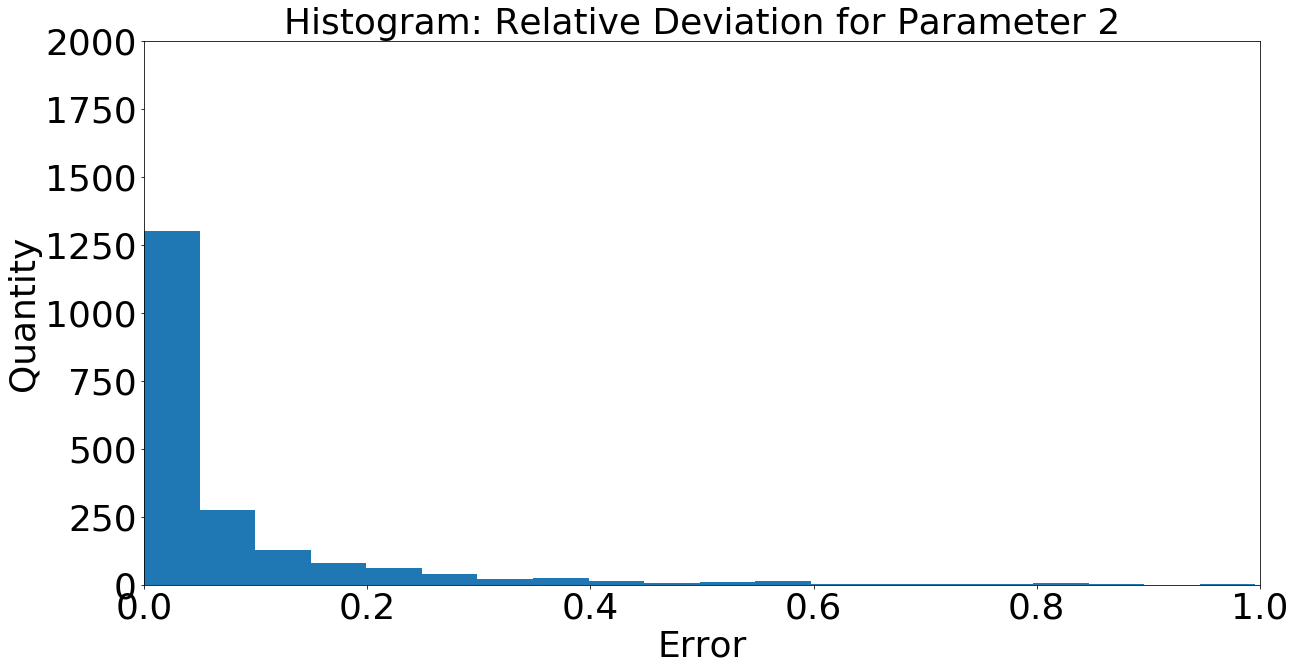}
    	\caption{Relative deviation of the prediction to the true parameter value for $p_2$ at optimization step $i=500.000$ for the test dataset with the objective $J_U$.}
    \end{minipage}
\end{figure}
We can recognize similar observations, when comparing training and test. Again, the red points seem to approximately fit the label points with high deviations for large parameter values. The deviation can be described by $\mu = 0.097$ and $\sigma = 0.127$ for the first parameter and $\mu = 0.076$ and $\sigma = 0.127$ for the second one.\\
Again, summarizing the results, we observe that the values for $\mu$ and $\sigma$ lie within an comparable range for the training as well as for the test data. Therefore, also minimizing the objective function $J_U$ results in a good generalizability for the neural network. Nevertheless, the prediction quality is slightly worse with mean relative deviation of around $8-10\%$ compared to objective $J_L$ with $4-8\%$.

\subsection{Robustness for Noisy Test Data}
For the experiments of the previous sections, we have always considered that the training and test data are clean or already de-noised. We have also seen the performance of a convolutional neural network for time series, being trained with a labelled and an unlabelled objective function. For these two experiments, the labelled approach is superior in terms of training and test error compared to the unlabelled approach. We now want to have a look at the following case: The model, for both, labelled and unlabelled learning, is trained using de-noised acceleration data. Therefore, the quality of the objective $J_U$ is ensured, using numerical integration to approximate the values for the corresponding velocity and state.\\
Now assume that for the test dataset, the acceleration is noisy, meaning for each $\left( \hat{\ddot{z}}, \hat{\ddot{y}} \right)\in \mathbb{R}^{N \times 2}$ within our test data, there is one corresponding Gaussian noise $\xi \in \mathbb{R}^N$ for $\ddot{z}$ and one corresponding Gaussian noise $\nu \in \mathbb{R}^N$ for $\ddot{y}$ such that the input to the neural network is given by 
\begin{equation}
\left( \bar{\ddot{z}}, \bar{\ddot{y}} \right) = \left(  \hat{\ddot{z}} + \xi, \hat{\ddot{y}} + \nu \right).
\end{equation}
We therefore try to compare which approach is more adequate to process noisy test data. Usually, neural networks can react very sensitive to data noise. Now, we test both trained networks, at the one hand the labelled and on the other hand the unlabelled approach and compare the performance for the noisy test data with $\xi_k, \nu_k \sim \mathcal{N}(0, 0.01)$ for all $k \in \{0,1,\hdots,N-1\}$.\\
\begin{figure}[!h]
	\centering
    \begin{minipage}[t]{7.5cm}
    	\centering
    	\includegraphics[width=7.5cm]{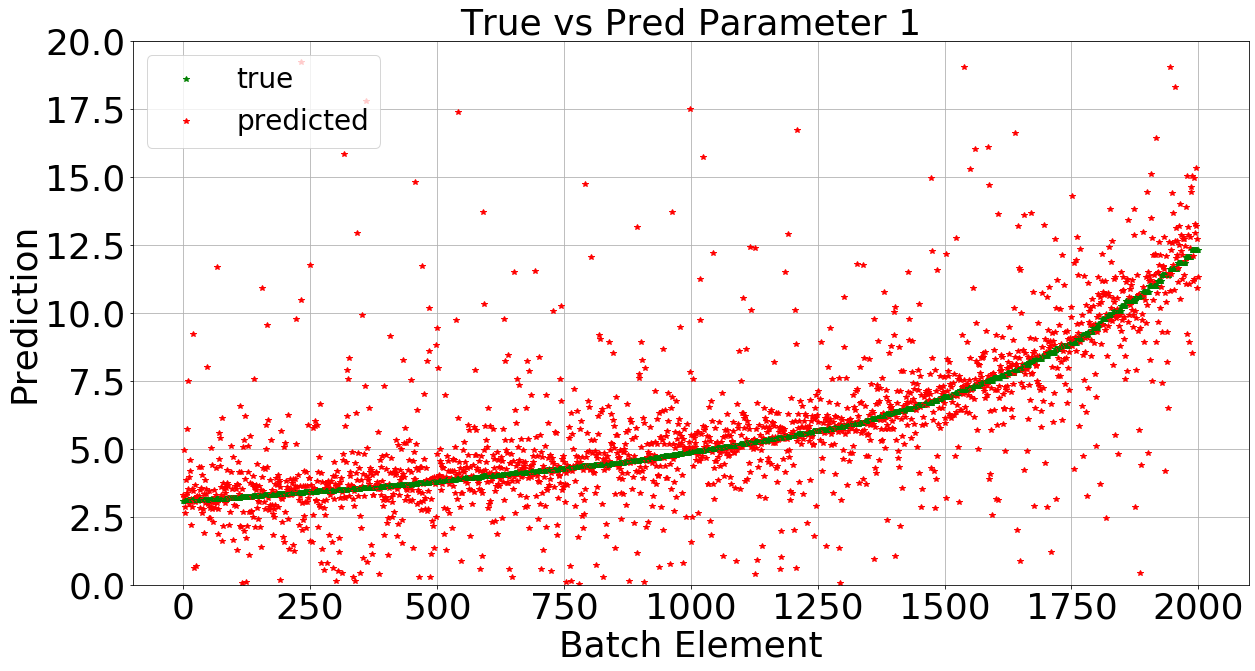}
    	\caption{Predictions (red) compared to true parameter values (green) for $p_1$ at optimization step $i=500.000$ for the test dataset with the objective $J_L$ and Gaussian noise, where $\sigma = 0.01$.}
    \end{minipage}
    \hspace{1cm}
    \begin{minipage}[t]{7.5cm}
    	\centering
    	\includegraphics[width=7.5cm]{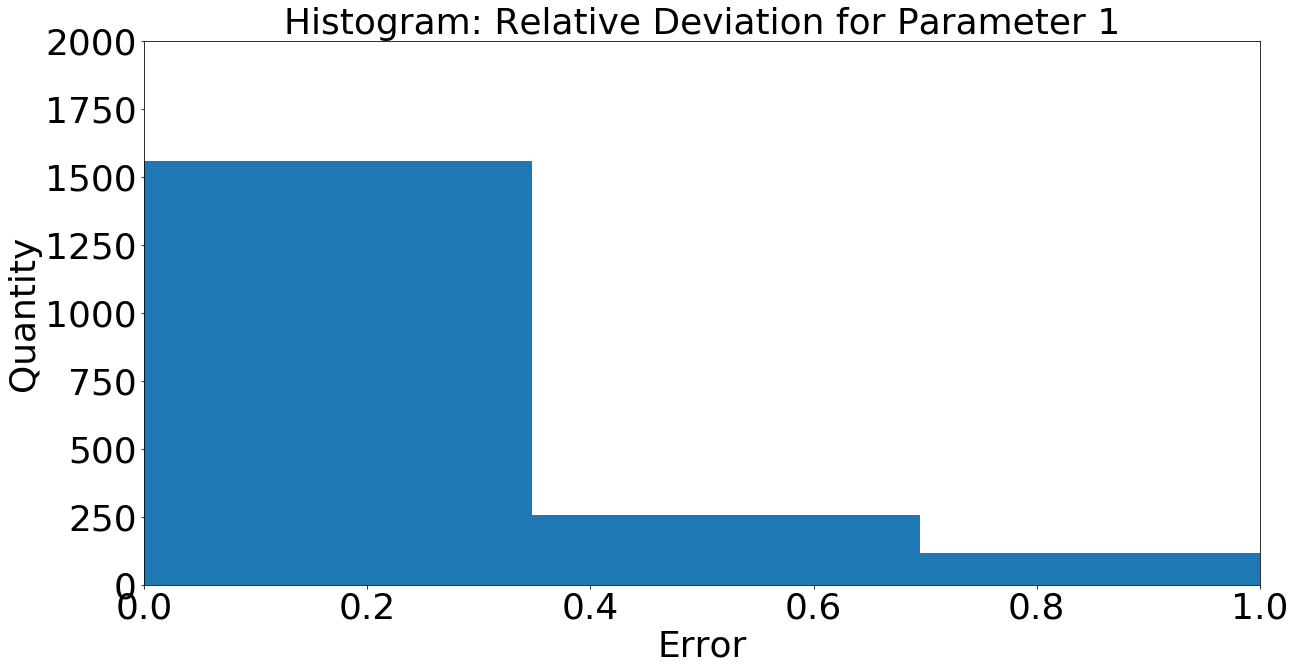}
    	\caption{Relative deviation of the prediction to the true parameter value for $p_1$ at optimization step $i=500.000$ for the test dataset with the objective $J_L$ and Gaussian noise, where $\sigma = 0.01$.}
    \end{minipage}
     
    \begin{minipage}[t]{7.5cm}
    	\centering
    	\includegraphics[width=7.5cm]{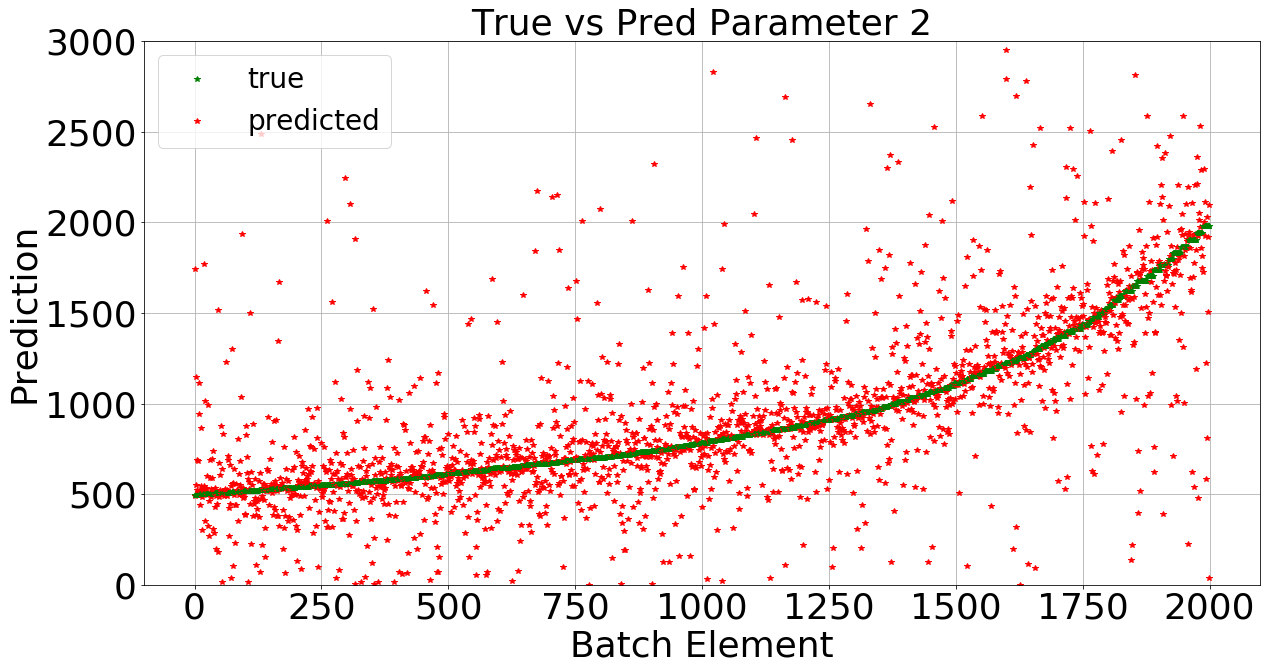}
    	\caption{Predictions (red) compared to true parameter values (green) for $p_2$ at optimization step $i=500.000$ for the test dataset with the objective $J_L$ and Gaussian noise, where $\sigma = 0.01$.}
    \end{minipage}
    \hspace{1cm}
    \begin{minipage}[t]{7.5cm}
    	\centering
    	\includegraphics[width=7.5cm]{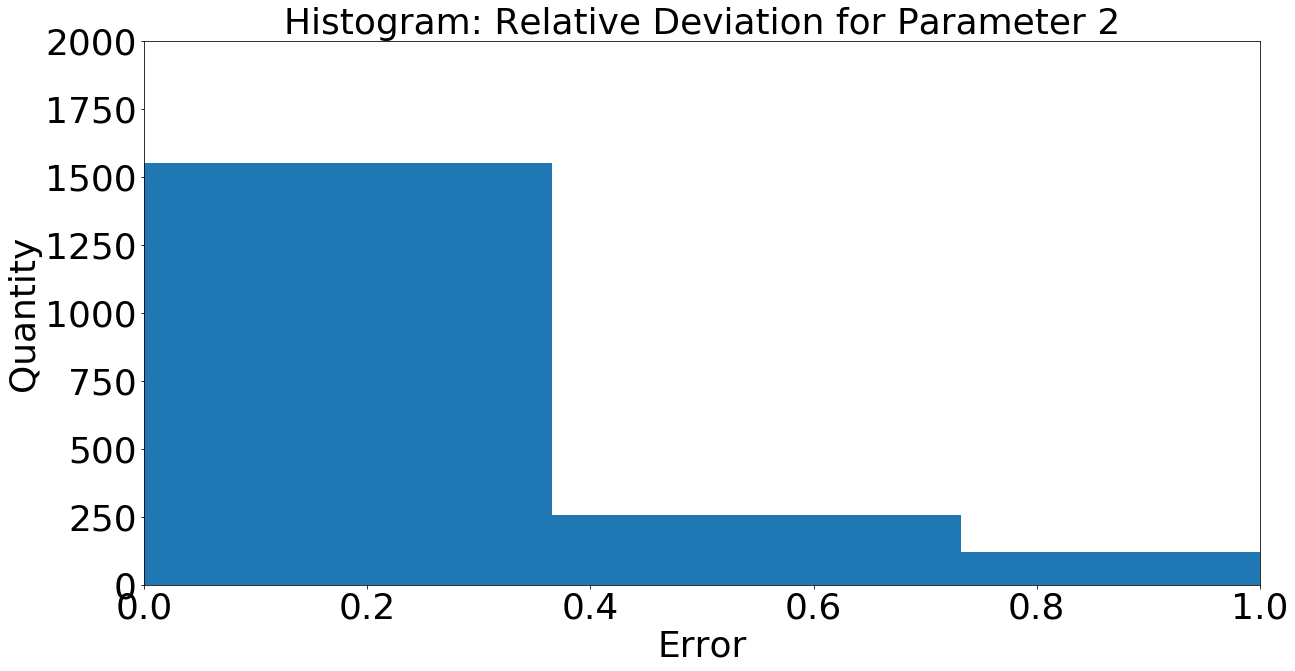}
    	\caption{Relative deviation of the prediction to the true parameter value for $p_2$ at optimization step $i=500.000$ for the test dataset with the objective $J_L$ and Gaussian noise, where $\sigma = 0.01$.}
    \end{minipage}
\end{figure}
As can be seen for the objective function $J_L$ and the test dataset (compare Fig.$22-25$), the output of the neural network is quite sensitive to the Gaussian noise added to the test dataset, as expected. We can recognize a decreasing performance of the prediction quality. Although the most part of the red point cloud is concentrated near the green labels, we can see that there are more values spread compared to the clean test data in Section $3.1$. As mentioned previously, we can recognize that the objective $J_L$ can give highly precise values for the training data: For both parameters $p_1$ and $p_2$ we get mean relative deviations of $4-8\%$. Contrarily, the performance for the noisy test dataset is much worse in this case: For both $p_1$ and $p_2$ the test deviation lies between $26-28\%$ throughout the training process. The generalizability for the convolutional neural network is for this specific case not given any more.\\
\begin{figure}[!h]
	\centering
    \begin{minipage}[t]{7.5cm}
    	\centering
    	\includegraphics[width=7.5cm]{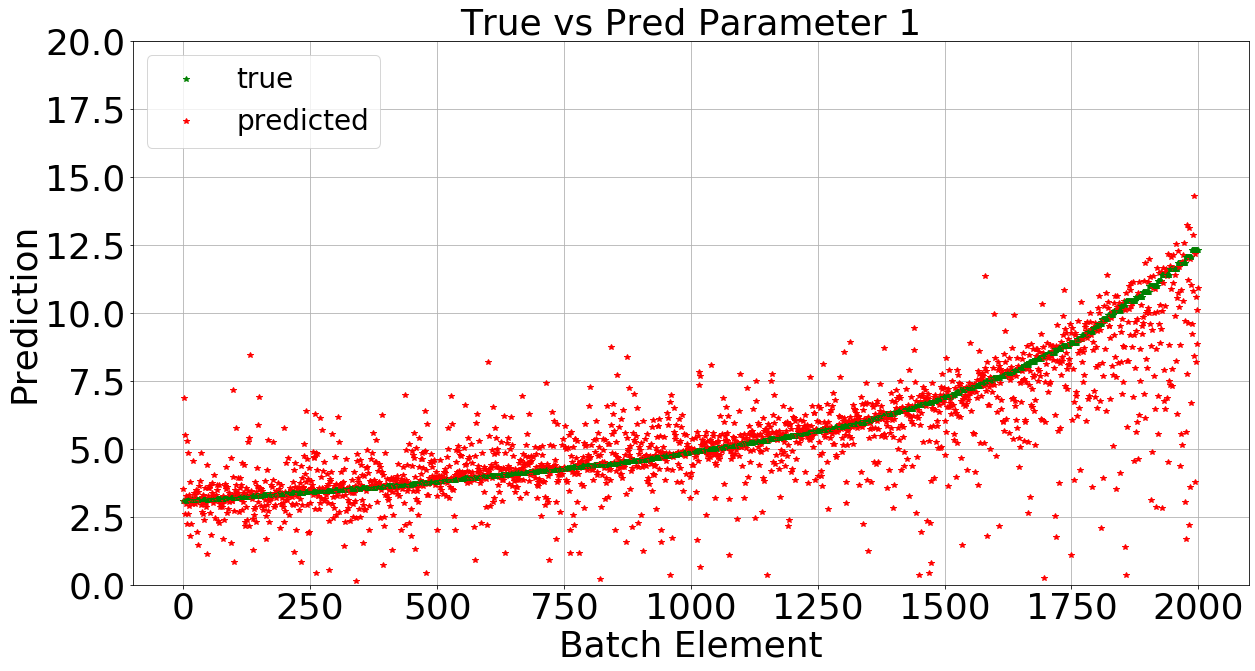}
    	\caption{Predictions (red) compared to true parameter values (green) for $p_1$ at optimization step $i=500.000$ for the test dataset with the objective $J_U$ and Gaussian noise, where $\sigma = 0.01$.}
    \end{minipage}
    \hspace{1cm}
    \begin{minipage}[t]{7.5cm}
    	\centering
    	\includegraphics[width=7.5cm]{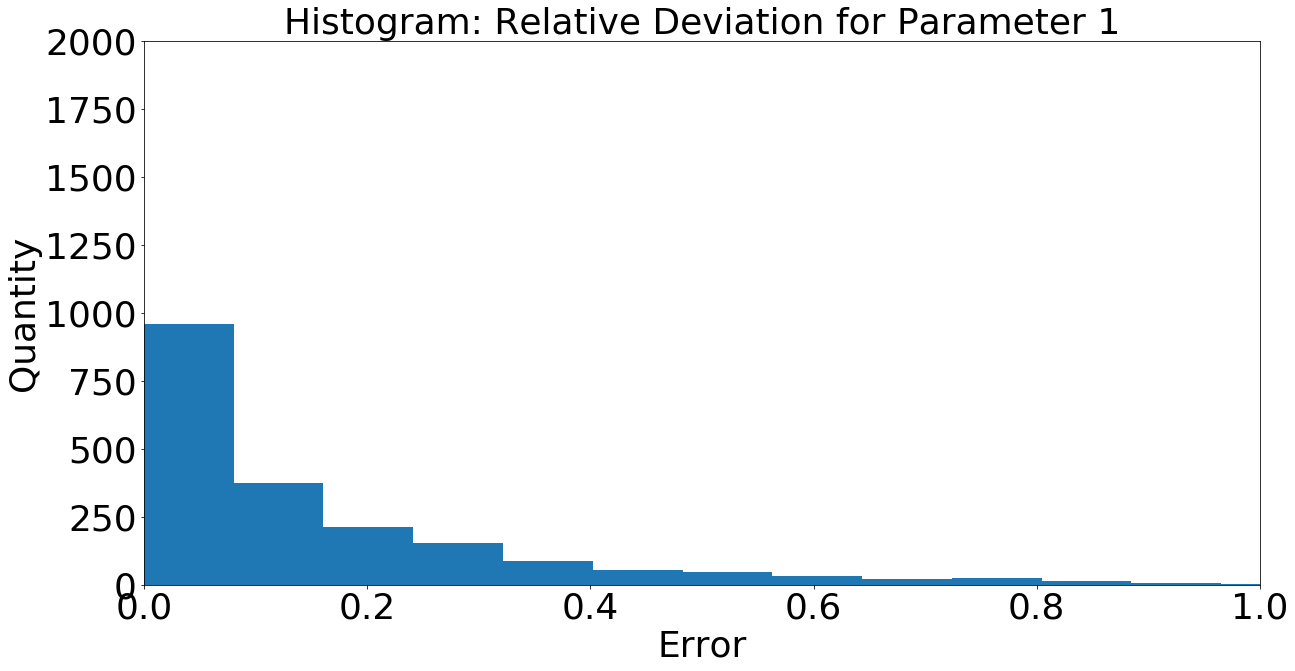}
    	\caption{Relative deviation of the prediction to the true parameter value for $p_1$ at optimization step $i=500.000$ for the test dataset with the objective $J_U$ and Gaussian noise, where $\sigma = 0.01$.}
    \end{minipage}
     
    \begin{minipage}[t]{7.5cm}
    	\centering
    	\includegraphics[width=7.5cm]{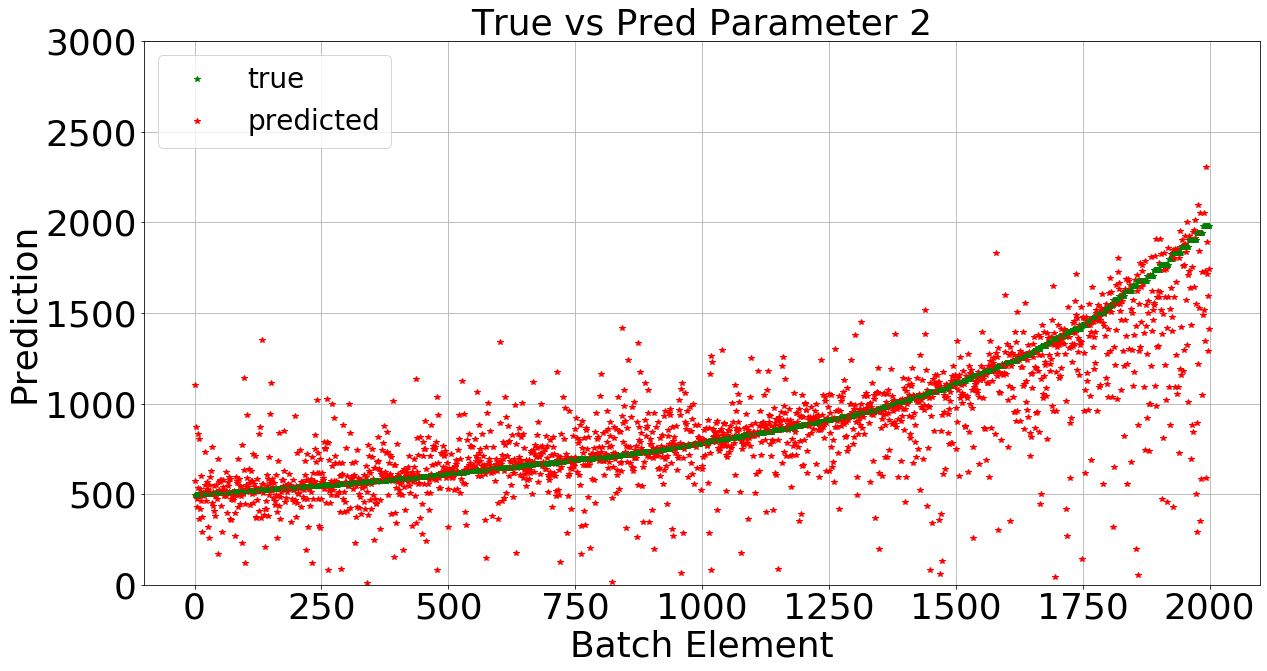}
    	\caption{Predictions (red) compared to true parameter values (green) for $p_2$ at optimization step $i=500.000$ for the test dataset with the objective $J_U$ and Gaussian noise, where $\sigma = 0.01$.}
    \end{minipage}
    \hspace{1cm}
    \begin{minipage}[t]{7.5cm}
    	\centering
    	\includegraphics[width=7.5cm]{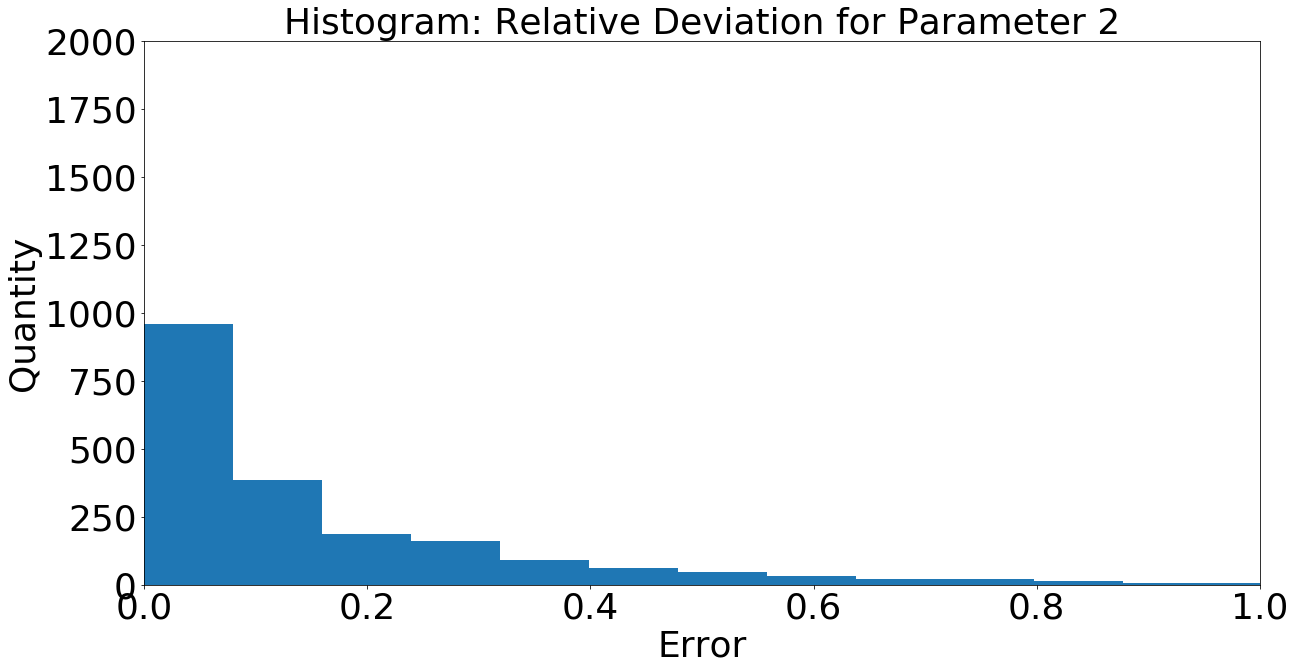}
    	\caption{Relative deviation of the prediction to the true parameter value for $p_2$ at optimization step $i=500.000$ for the test dataset with the objective $J_U$ and Gaussian noise, where $\sigma = 0.01$.}
    \end{minipage}
\end{figure}
There are significantly better results in predicting the values of the test data when the unlabelled objective $J_U$ is used for the training procedure. For the training dataset, the unlabelled objective function results in an adjustment of the network's parameters to predict with a relative deviation of $8-10\%$. The noisy test data results in a relative deviation of around $16\%$. Consequently, the performance for the training data is very close to the performance of the test data and therefore the second objective is more robust compared to the labelled one.\\
\begin{figure}[!h]
	\centering
    \begin{minipage}[t]{7.5cm}
    	\centering
    	\includegraphics[width=7.5cm]{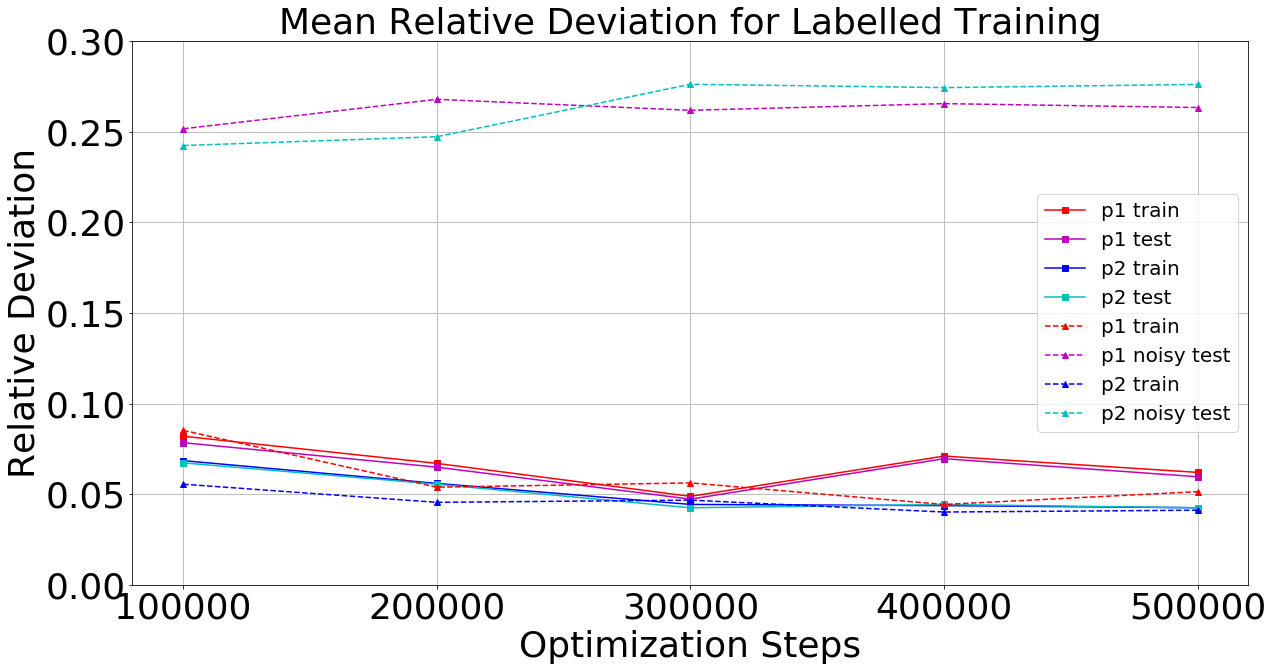}
    	\caption{Mean relative deviation of $p_1$ and $p_2$ for each $100000$ steps for two optimization processes using $J_L$: $(1)$ optimization process using de-noised training and de-noised test data. Mean deviation is shown for $p_1$ training (red line), $p_1$ test (violet line), $p_2$ training (blue line) and $p_2$ test (cyan line). $(2)$ optimization process using de-noised training and noisy test data. Mean deviation is shown for $p_1$ training (red dashed line), $p_1$ test (violet dashed line), $p_2$ training (blue dashed lined) and $p_2$ (cyan dashed line).}
    \end{minipage}
    \hspace{1cm}
    \begin{minipage}[t]{7.5cm}
    	\centering
    	\includegraphics[width=7.5cm]{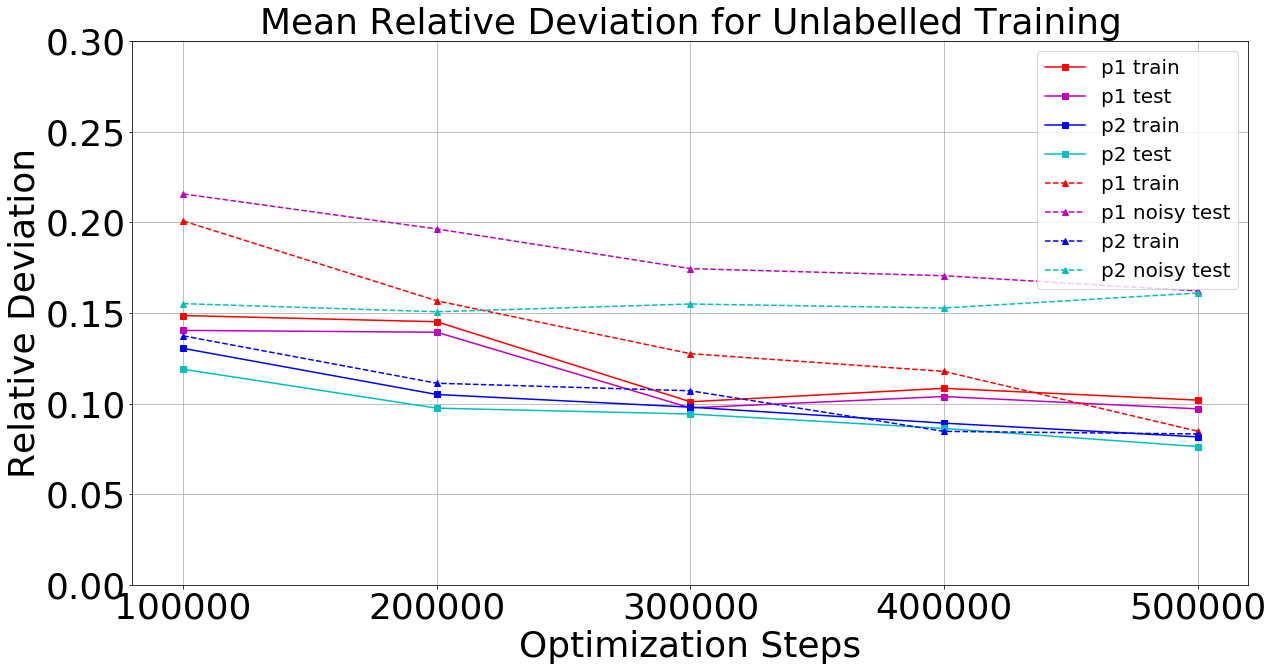}
    	\caption{Mean relative deviation of $p_1$ and $p_2$ for each $100000$ steps for two optimization processes using $J_U$: $(1)$ optimization process using de-noised training and de-noised test data. Mean deviation is shown for $p_1$ training (red line), $p_1$ test (violet line), $p_2$ training (blue line) and $p_2$ test (cyan line). $(2)$ optimization process using de-noised training and noisy test data. Mean deviation is shown for $p_1$ training (red dashed line), $p_1$ test (violet dashed line), $p_2$ training (blue dashed lined) and $p_2$ (cyan dashed line).}
    \end{minipage}
     
    \begin{minipage}[t]{7.5cm}
    	\centering
    	\includegraphics[width=7.5cm]{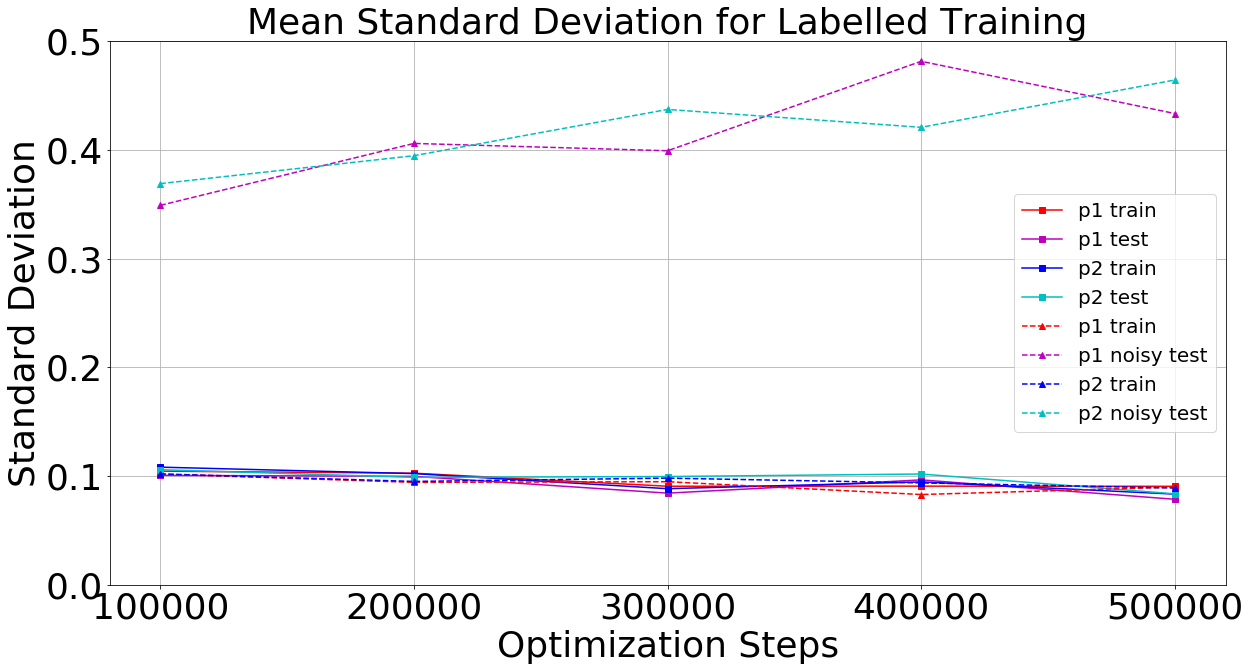}
    	\caption{Mean standard deviation of $p_1$ and $p_2$ for each $100000$ steps for two optimization processes using $J_L$: $(1)$ optimization process using de-noised training and de-noised test data. Mean deviation is shown for $p_1$ training (red line), $p_1$ test (violet line), $p_2$ training (blue line) and $p_2$ test (cyan line). $(2)$ optimization process using de-noised training and noisy test data. Mean deviation is shown for $p_1$ training (red dashed line), $p_1$ test (violet dashed line), $p_2$ training (blue dashed lined) and $p_2$ (cyan dashed line).}
    \end{minipage}
    \hspace{1cm}
    \begin{minipage}[t]{7.5cm}
    	\centering
    	\includegraphics[width=7.5cm]{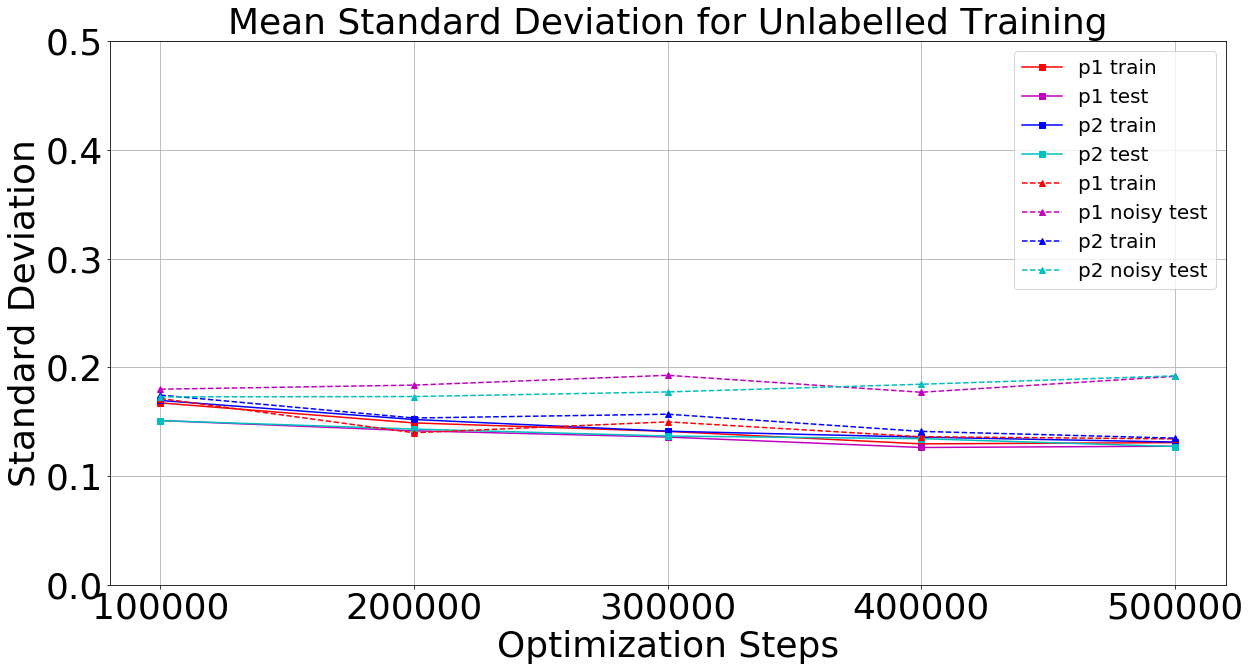}
    	\caption{Mean standard deviation of $p_1$ and $p_2$ for each $100000$ steps for two optimization processes using $J_U$: $(1)$ optimization process using de-noised training and de-noised test data. Mean deviation is shown for $p_1$ training (red line), $p_1$ test (violet line), $p_2$ training (blue line) and $p_2$ test (cyan line). $(2)$ optimization process using de-noised training and noisy test data. Mean deviation is shown for $p_1$ training (red dashed line), $p_1$ test (violet dashed line), $p_2$ training (blue dashed lined) and $p_2$ (cyan dashed line).}
    \end{minipage}
\end{figure}
The robustness of the second objective compared to the first one becomes obvious in Fig.$30-33$. In Fig.$30$, the mean relative deviation, for both parameter $p_1$ and parameter $p_2$, are shown at every $i = \lambda \cdot 100000$ iteration steps with $\lambda \in \{1,2,3,4,5\}$ for the training as well as for the test dataset using objective $J_L$. The straight lines represents the evaluation for the optimization process using the de-noised test data, the dashed line for the noisy test data.\\
In an analogous way, in Fig. $32$, the results for the mean standard deviation are shown, again using objective $J_L$. As can be seen in both Fig. $30$ and Fig. $32$, the performance lines are close to each other using the de-noised test data approach. When considering the noisy test data for parameter $p_1$ (violet dashed line) and $p_2$ (cyan dashed line) there is a large gap compared to the training deviation.\\
Therefore, we again see that the approach using $J_L$ works well as far as generalizability for de-noised test data is concerned but fails for noisy samples.\\
In contrast to that, we can have a look at the mean relative deviation for $J_U$ in Fig.$31$ and the corresponding mean standard deviation in Fig.$33$. As can be seen, the dashed lines and the straight lines are more close to each other with an absolute lower error for the noisy test data and the unlabelled approach.\\
Therefore, we can conclude that the second objective is more robust for parameter estimation using noisy test data in a convolutional neural network.\\

\section{Conclusion}
We have discussed system identification of approximative vehicle model's parameters, given by a non-homogeneous system of second order ordinary differential equations. Therefore, acceleration data of the system's masses has been generated, using a symplectic Euler scheme for numerical integration of the differential equations and different random generations of non-homogeneous components and masses. A one-dimensional convolutional network has been applied as a model to predict the parameters of the underlying differential equation based on acceleration profiles. The training has been carried out with respect to two different objective functions, one of which used the true values of the equation's parameters and the second of which encodes the reproduction of the input data. It has been shown that for clean test data, both objectives result in acceptable performances on the test data, where the first objective slightly outperforms the second one. In contrast, if test samples with additive Gaussian noise are processed, the network trained with the second objective is significantly more robust against noise.\\
It is worth doing further investigations in this field, to find the root cause of the results. Simple mathematical models, that show similar results following the approaches in this work, could be taken into consideration to get a deeper understanding of this robustness effect.\\

\newpage

\bibliographystyle{unsrt}  


\end{document}